\documentclass[lettersize,journal]{IEEEtran}
\usepackage{amsmath,amsfonts,amssymb}
\usepackage{algorithmic}
\usepackage{algorithm}
\usepackage{array}
\usepackage[caption=false,font=normalsize,labelfont=sf,textfont=sf]{subfig}
\usepackage{textcomp}
\usepackage{stfloats}
\usepackage{hyperref}
\usepackage{verbatim}
\usepackage{graphicx}
\usepackage{cite}
\usepackage{graphics} 
\usepackage{epsfig} 
\usepackage{cleveref}
\usepackage{booktabs}
\usepackage{siunitx}
\usepackage{multirow}
\usepackage{bm}
\usepackage{color,soul} 

\usepackage{xcolor}
\newcommand{\rv}[1]{\textcolor{black}{#1}}

\newcommand{\ind}{\textbf{1}}

\begin{document}

\title{Interactive Autonomous Navigation with \\ Internal State Inference and Interactivity Estimation
}

\author{Jiachen Li, David Isele, Kanghoon Lee, Jinkyoo Park, Kikuo Fujimura, and Mykel J. Kochenderfer 
	\thanks{J. Li, J. Park, and M. J. Kochenderfer are with the Stanford Intelligent Systems Laboratory (SISL), Stanford University, CA, USA. K. Lee and J. Park are with the Korea Advanced Institute of Science and Technology, Korea. {\tt\small \{jiachen\_li,jkpark11,mykel\}@stanford.edu; leehoon@kaist.ac.kr}.}
	\thanks{D. Isele and K. Fujimura are with the Honda Research Institute USA, CA, USA {\tt\small \{disele, kfujimura\}@honda-ri.com}.}
}



\maketitle

\begin{abstract}

Deep reinforcement learning (DRL) provides a promising way for intelligent agents (e.g., autonomous vehicles) to learn to navigate complex scenarios. 
However, DRL with neural networks as function approximators is typically considered a black box with little explainability and often suffers from suboptimal performance, especially for autonomous navigation in highly interactive multi-agent environments.
To address these issues, we propose three auxiliary tasks with spatio-temporal relational reasoning and integrate them into the standard DRL framework, which improves the decision making performance and provides explainable intermediate indicators.
We propose to explicitly infer the internal states (i.e., traits and intentions) of surrounding agents (e.g., human drivers) as well as to predict their future trajectories in the situations with and without the ego agent through counterfactual reasoning. These auxiliary tasks provide additional supervision signals to infer the behavior patterns of other interactive agents.
Multiple variants of framework integration strategies are compared.
We also employ a spatio-temporal graph neural network to encode relations between dynamic entities, which enhances both internal state inference and decision making of the ego agent.
Moreover, we propose an interactivity estimation mechanism based on the difference between predicted trajectories in these two situations, which indicates the degree of influence of the ego agent on other agents.
To validate the proposed method, we design an intersection driving simulator based on the Intelligent Intersection Driver Model (IIDM) that simulates vehicles and pedestrians. 
Our approach achieves robust and state-of-the-art performance in terms of standard evaluation metrics and provides explainable intermediate indicators (i.e., internal states, and interactivity scores) for decision making. 
\end{abstract}


\section{Introduction}

\begin{figure}
	\centering
	\includegraphics[width=\columnwidth]{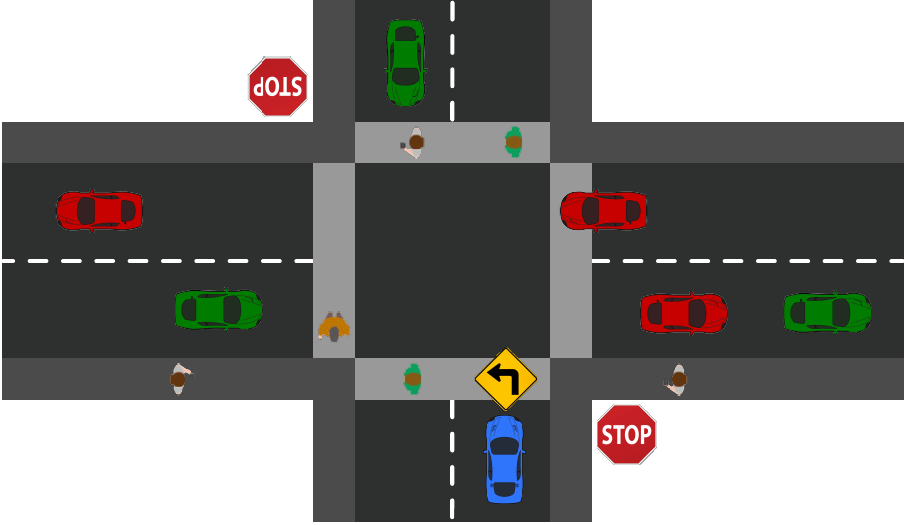}
	\caption{An illustrative partially controlled intersection with two-way stop signs. The ego vehicle (blue) is trying to turn left and merge into the horizontal lane without colliding with crossing pedestrians, aggressive (red), or conservative (green) vehicles. Best viewed in color.}
	\label{fig:teaser}
\end{figure}

Controlling autonomous vehicles in urban traffic scenarios (e.g., intersections) is a challenging sequential decision making problem that must consider the complex interactions among heterogeneous traffic participants (e.g., human-driven vehicles, pedestrians) in dynamic environments.
Consider a partially controlled intersection where an autonomous vehicle tries to turn left from a lane with a stop sign and the crossing traffic does not stop except when there are crossing pedestrians ahead (see Fig. \ref{fig:teaser}). 
On one hand, autonomous vehicles should not ignore the oncoming/crossing traffic and turn aggressively, which may lead to a collision. 
On the other hand, it should not be overly conservative, which can hurt efficiency.
In such a scenario, human drivers can reason about the relations between interactive entities, recognize other agents' intentions, and infer how their actions will affect the behavior of others on the road, allowing them to negotiate right of way and drive safely and efficiently.

Human drivers are internally heterogeneous in terms of both trait and intention \cite{zhu2019typical,brown2020taxonomy,ma2021reinforcement,bae2022lane}. Conservative drivers tend to yield to other traffic participants during interactions, keep a larger distance from their leading vehicles, and maintain a lower desired speed; aggressive drivers are the opposite.
To increase driving efficiency while maintaining safety, autonomous vehicles need to accurately infer the internal states of others, including traits (i.e., conservative/aggressive) and intentions (i.e., yield/not yield).
Besides these high-level cues, accurate opponent modeling in the form of multi-agent trajectory prediction provides additional cues for safe and efficient decision making.
Ma et al. \cite{ma2021reinforcement} incorporated trait inference as an auxiliary task into a reinforcement learning framework and achieved better performance than standard reinforcement learning.
In this work, we use additional supervision from driver intention recognition and trajectory prediction to enhance performance.
\rv{We validate the hypothesis that modeling human internal states explicitly improves the decision making performance and the inferred internal states can serve as explainable indicators.}

To negotiate with other traffic participants, autonomous vehicles must infer to what extent they can influence the behavior of others.
Not all the agents in the scene have strong interactions with the ego vehicle.
In Fig. \ref{fig:teaser}, the ego vehicle only needs to negotiate with the red vehicle in the upper right corner which has conflicts in their future paths. Although the other two green vehicles also approach the intersection, they need to yield to the crossing pedestrians and thus will not influence the ego agent's actions in the short-term future, and vice versa. 
Existing approaches usually adopt soft attention mechanisms to learn the importance weights of different agents \cite{vemula2018social,huang2019stgat,kosaraju2019social}. However, these techniques may assign small weights to important objects or large weights to irrelevant ones \cite{vemula2018social,li2021rain}, which can mislead the decision making of autonomous vehicles, especially in dense scenarios.  

In this paper, we propose a mechanism to estimate the interactivity between the ego vehicle and each surrounding agent by using the difference between the predicted trajectory distributions of the agent under the situations with and without the existence of the ego vehicle as a quantitative indicator. 
This difference can also be treated as a quantitative \textit{degree of influence} that the ego vehicle can have on a certain agent, which is called ``interactivity score''.
The interactivity scores are also used to weigh the prediction errors in the loss function, which encourages the model to generate more accurate trajectories for the agents that may have stronger interactions with the ego vehicle.
Note that the ego vehicle always exists in both training and testing environments, thus predicting the future behaviors of other agents without the existence of the ego vehicle is a \textit{counterfactual reasoning} problem.
We propose to use a prediction model pre-trained with the trajectory data collected in the environments without the ego vehicle to generate counterfactual predictions in the formal training process. The weights of this prediction model are fixed without further updates.

Autonomous vehicles need the ability to understand and reason about the interactions between dynamic entities by modeling their spatio-temporal relations. It is natural to represent a multi-agent system as a graph, where node attributes encode the information of agents and edge attributes encode their relations or interactions. Recently, graph neural networks have been widely adopted to capture relational features and model interactions between multiple entities \cite{li2020evolvegraph, kipf2018neural, li2021grin, girgis2021latent,huang2019stgat}.
In this work, we employ a spatio-temporal graph neural network as the basis model for spatio-temporal relational reasoning.

Among sequential decision making approaches, deep reinforcement learning has been widely studied for autonomous navigation in complex scenarios due to its high capability of learning flexible representations and policies \cite{aradi2020survey,wei2021autonomous}.
Despite its promising performance, the explainability of these methods still remains underexplored, which is crucial for safety-critical applications \cite{omeiza2021explanations}.
Some approaches learn an attention map on the perceived images to indicate the salient areas, which mimics the gaze of human drivers \cite{martin2018dynamics,li2021spatio}. However, the learned attention map can be misleading because soft attention may assign unreasonable weights \cite{huang2019stgat,li2021rain}. 
Also, the learned attention weights are associated with image pixels instead of agents, which makes it difficult to provide explanations at the object level.
Other methods apply soft attention mechanisms to object-level entities instead, yet this limitation still applies.
In this work, we enhance the transparency of the decision making process with both internal state inference and interactivity estimation, which provides intermediate auxiliary features to indicate how the model infers and reasons about the agents' interactive behaviors.

The main contributions of this paper are as follows:
\begin{itemize}
	\item We propose a deep reinforcement learning approach for interactive autonomous navigation with three auxiliary tasks: internal state inference, trajectory prediction, and interactivity estimation. Multiple variants of framework architectures are compared empirically.
	\item \rv{The auxiliary tasks not only improve the decision making performance but also enhance the transparency of the proposed framework by inferring explainable intermediate features of surrounding agents. In particular, we propose an explainable technique to estimate interactivity scores based on the ego agent's degree of influence on surrounding agents through counterfactual reasoning.}
    \item \rv{We design a four-way partially controlled intersection environment that simulates challenging traffic scenarios with interactive vehicles and crossing pedestrians, which is used to validate our approach and can serve as a novel benchmark for future research.}
	\item Our approach demonstrates superior performance compared to baselines in terms of completion rate, collision rate, driving efficiency in a complex intersection as well as stronger robustness to out-of-distribution scenarios.
\end{itemize}

This paper builds upon our previous work \cite{ma2021reinforcement} in several important ways.
First, the internal state in our earlier work only contains the traits of human drivers while we additionally infer their intentions in this work to model different internal aspects and randomness in human behaviors.
Second, we provide a more comprehensive discussion and comparison between different variants of framework architectures to incorporate the internal state inference.
Third, we propose two additional auxiliary tasks (i.e., trajectory prediction, and interactivity estimation) into the reinforcement learning framework, which improves the decision making performance and enhances the transparency of the proposed framework.
Finally, we design a more challenging intersection driving simulator with crossing pedestrians based on IIDM to validate our approach.

The remainder of the paper is organized as follows. 
Section II provides a concise summary of the related work.
Section III introduces basic background knowledge related to the proposed approach. 
Section IV introduces the intersection driving simulation based on the IIDM, which is adopted as the training and testing environment in our experiments.
Section V presents the problem formulation for the autonomous navigation task.  
Sections VI and VII introduce the details of the proposed method.
Section VIII presents the experimental settings, quantitative and qualitative results, and analysis.
Finally, Section IX concludes the paper and discusses the impacts and potential limitations of our method. 

\section{Related Work}

\subsection{Autonomous Vehicle Navigation}
Decision making and motion planning for autonomous vehicles have been widely studied \cite{wei2021autonomous}.
\rv{Earlier approaches use control theory, optimization, and classical artificial intelligence techniques to plan a future trajectory for autonomous vehicles \cite{wilfong1990motion,ferguson2008motion,anderson2010optimal,wei2011point,frazzoli2000robust,waydo2003vehicle,zhang2017optimal}.} These methods can effectively handle autonomous navigation in simple environments.
Recently, many research efforts have been devoted to designing learning-based approaches for autonomous driving. 
To handle more complex environments with dynamic, interactive agents, some approaches adopt a game-theoretic perspective to model interactions \cite{fisac2019hierarchical,wang2021game,hang2020human,li2023game}.
However, it is difficult to extend these methods to large-scale interacting systems with many entities.
Imitation learning can be used to train an autonomous vehicle to navigate in an end-to-end manner \cite{pan2018agile}. However, imitation learning struggles to generalize well to out-of-distribution scenarios, and it demands the collection of a large set of expert demonstrations.

Deep reinforcement learning has been widely adopted to solve sequential decision making problems in modern intelligent systems \cite{franccois2018introduction} and can be applied to autonomous driving \cite{aradi2020survey}.
Some approaches take in raw sensor measurements (e.g., RGB images, point cloud) and outputs control commands (e.g., acceleration, steering angle) in an end-to-end manner, modeling the interactions between dynamic entities implicitly \cite{codevilla2018end,chitta2021neat,amini2020learning,zhang2021end}. 
Although these approaches may achieve satisfactory performance, it is difficult to interpret these methods and understand the underlying interactions between interactive agents.
\rv{Another category of approaches uses the low-dimensional state information (e.g., position, velocity) of agents provided by upstream perception modules, which put more emphasis on behavior modeling and multi-agent coordination interaction \cite{ma2021reinforcement,lee2023robust,saxena2020driving,isele2018navigating,guan2022integrated}.} However, the information provided by the perception module could be noisy.
Our method falls into this category, and we propose to learn auxiliary tasks that help reinforcement learning and generate meaningful intermediate indicators that enable explainable autonomous navigation.

\subsection{Interactive Decision Making}
Depending on the scope of controlled agents, different formulations and algorithms can be employed. 
If all the agents are co-learning agents whose policies can be updated, then the game-theoretic formulation and multi-agent reinforcement learning can be employed \cite{hang2020human,hang2020integrated,li2016hierarchical}. These approaches generally model autonomous navigation as a non-cooperative game and use the Nash equilibrium and Stackelberg game theory to produce human-like behaviors. 
When we treat other agents as part of the environment reacting to the ego agent, single-agent-based control with opponent modeling can be employed to control the ego agent \cite{liu2021deep}.
In these approaches, modeling the opponents' interactive behaviors is essential for deriving a tractable control policy. 
Without opponent modeling, the environment is non-stationary, which makes the learned policy for the ego vehicle unreliable. 
Our work falls into the second category and focuses on deriving the control policy of the ego vehicle in a reinforcement learning framework with opponent modeling as auxiliary supervised learning tasks to extract essential factors for decision making.

\subsection{Behavior Modeling and Trajectory Prediction}
Behavior modeling of other agents can be incorporated into the environment dynamics, transforming the multi-agent problem into a single-agent control problem \cite{rabinowitz2018machine,hernandez2019agent,papoudakis2021agent}. 
Recent works apply latent factor modeling to extract stochastic internal factors of agents \cite{ma2021reinforcement,wang2021interpretable,xie2020learning,zintgraf2021deep}. 
These studies use the extracted hidden representations as input for the control policy of the ego vehicle while interpreting them as the intentions, goals, or the strategy of other agents. 
Ma et al. \cite{ma2021reinforcement} propose a framework to explicitly infer the internal state of agents and integrate the module into the reinforcement learning framework for driving policy learning.
Wang et al. \cite{wang2021interpretable} propose a latent model of vehicle behaviors at highway on-ramps to produce interpretable behaviors. 
Xie et al. \cite{xie2020learning} propose a reinforcement learning framework with latent representation learning of other agents' policies.
In this work, instead of only inferring high-level internal states (e.g., traits, intentions) of other agents, we also incorporate trajectory prediction explicitly into the decision making framework.

Accurate trajectory prediction of other traffic participants in a multi-agent environment is an essential step for controlling the ego vehicle. 
Many methods have been proposed to create more expressive models to capture the inherent complexity of multi-agent behaviors while taking into account the latent goal/intention of interactive agents \cite{li2019generic, gu2021densetnt,fan2021intention, girase2021loki,zhou2022grouptron,li2023ped}. 
Recently, graph neural networks combined with latent variable modeling have been widely applied to predict the future trajectories of multiple agents while considering their latent relations \cite{li2020evolvegraph, kipf2018neural, li2021grin, girgis2021latent}. 
These works only focus on improving the trajectory prediction accuracy without validating its actual effectiveness in downstream tasks.
In this work, we integrate prediction into the decision making framework and demonstrate its effectiveness. 

\subsection{Counterfactual Reasoning}

Humans often create counterfactual alternatives to reality to answer  ``what if'' questions by thinking about how things could have turned out differently if they make a different action \cite{hoch1985counterfactual}.
In a multi-agent setting, counterfactual reasoning is often adopted to facilitate social interactions. 
Jaques et al. \cite{jaques2019social} propose to use social influence as an intrinsic reward to encourage cooperative agents to learn to actively influence the other agents' policies to obtain a larger expected return. 
\rv{Tolstaya et al. \cite{tolstaya2021identifying} and Khandelwal et al. \cite{khandelwal2020if} propose a conditional behavior prediction method by forecasting the future behavior of a target agent conditioned on a counterfactual future behavior of the query agent.}
In this work, we propose counterfactual prediction by removing the ego agent from the scene and calculating the difference in the future behavior of a target agent, which is used to estimate the interactivity between the ego agent and the target agent for ego decision making.

\section{Preliminaries}

\subsection{Partially Observable Markov Decision Process (POMDP)}

A Markov decision process (MDP) is typically used to describe a discrete-time stochastic sequential decision making process where an agent interacts with the environment.
Formally, an MDP is specified by the tuple $(S, A, T, R, \gamma, \rho_0)$ where $S$ and $A$ denote the state and action space, $T$ denotes the transition model, $R$ denotes the reward, $\gamma \in [0,1]$ denotes the discount factor, and $\rho_0$ denotes the initial state distribution.
A partially observable Markov decision process (POMDP) is a generalization of a MDP, where the agent cannot directly observe the complete state. An additional observation function $\Omega$ is needed to map a state $s \in S$ to an observation $o \in O$ where $O$ denotes the observation space.
Formally, a POMDP is specified by the tuple $(S, A, T, R, \Omega, O, \gamma, \rho_0)$.
Unlike the policy function in MDP which maps states to actions, the policy of a POMDP maps the historical observations (or belief states) to actions.
The objective is to find a policy $\pi$ that maximizes the expected return
\begin{align}
	\pi^* = \arg \max_\pi \mathbb{E}_{s_0, a_0, o_0,...} \sum_{t=0}^{\infty} \gamma^t R(s_t, a_t),
\end{align}
where $s_0 \sim \rho_0(s_0)$, $a_t \sim \pi(a_t \mid o_{1:t})$, $o_t \sim \Omega(o_t \mid s_t)$, $s_{t+1} \sim T(s_{t+1} \mid s_t, a_t)$, and $t$ denotes the index of time steps.

\subsection{Policy Optimization}

Policy gradient methods are widely used to learn optimal policies by optimizing the policy parameters directly \cite{williams1992simple,schulman2015trust,schulman2017proximal}. 
The traditional REINFORCE algorithm \cite{williams1992simple} provides an unbiased gradient estimator with the objective $L^\text{PG}(\theta)=\hat{\mathbb{E}}[\log \pi_\theta (a \mid s)\hat{A}]$, where $\hat{A}$ is the estimated advantage.
For a POMDP, we use the observation and the hidden state of the policy instead of the state $s$.
Recently, PPO \cite{schulman2017proximal} has been a widely used policy optimization algorithm due to its simplicity and stable training performance, in which a clipped surrogate objective is maximized
\begin{align}
	L^\text{PPO}(\theta) = \hat{\mathbb{E}}[\min (&r(\theta)\hat{A}, \text{clip}(r(\theta), 1-\epsilon, 1+\epsilon)\hat{A})],\\
	&r(\theta) = \frac{\pi_\theta(a \mid s)}{\pi_{\theta'}(a \mid s)},
\end{align}
where $\theta'$ denotes the parameters of the old policy that is used to collect experiences, and $\epsilon$ denotes the clipping threshold.

\subsection{Graph Neural Networks}

The graph neural network is a class of deep learning models that can be applied to process the information on graph-structured data. 
A specific design of the message passing mechanism naturally incorporates certain relational inductive biases into the model.
Most graphs are attributed (e.g., node attributes, edge attributes) in the context of graph neural networks. 
Battaglia et al. \cite{battaglia2018relational} provide a comprehensive introduction to graph neural networks.
Generally, there are two basic GNN operations in graph representation learning: edge update and node update.
More formally, we denote the graph with $N$ nodes as $\mathcal{G}=\{\mathcal{V},\mathcal{E}\}$, where $\mathcal{V}=\{v_i \mid i \in \{1,\ldots,N\}\}$ is a set of node attributes and $\mathcal{E}=\{e_{ij} \mid i,j \in \{1,\ldots,N\}\}$ is a set of edge attributes. Then, the two update operations are
\begin{align}
	e'_{ij} = \phi^e(e_{ij}, v_i, v_j), \
	\bar{e}'_i = f^{e\rightarrow v}(E'_i), \
	v'_i = \phi^v(\bar{e}'_i, v_i),
\end{align}
where $E'_i=\{e'_{ij} \mid j\in \mathcal{N}^i\}$, $E'=\bigcup_i E'_i$, $V'=\{v'_i \mid i=1,\ldots,n\}$, and $\mathcal{N}^i$ is the direct neighbors of node $i$.
We denote $\phi^e(\cdot)$ and $\phi^v(\cdot)$ as deep neural networks.
We denote $f^{e\rightarrow v}(\cdot)$ as an arbitrary aggregation function with the property of permutation invariance.

\begin{table}[!tbp]
	\centering
	\caption{Notations of variables and parameters in IDM.}
	\begin{tabular}{@{}lll@{}}
		\toprule
		Symbol & Description & Unit/Value \\
		\midrule
		$s$ & distance from the leading vehicle & \SI{}{m} \\
		$v$ & longitudinal velocity & \SI{}{m/s} \\
		$\Delta v$ & approaching rate & \SI{}{m/s} \\
		\midrule
		$\delta$ & free-drive exponent & 4 \\
		$s_0$ & minimum distance from leading vehicle & \rv{\SI{4.5}{\meter}--\SI{9.0}{\meter}} \\
		$v^*$ & desired velocity & \rv{\SI{8.4}{m/s}--\SI{9.0}{m/s}} \\
		$T$ & safety time gap & \SI{1.5}{\second} \\
		$a_\text{max}$ & maximum acceleration & \SI{3.0}{m/s^2} \\
		$b_\text{comf}$ & comfortable braking deceleration & \SI{2.0}{m/s^2} \\
		\bottomrule
	\end{tabular}
	\label{tab:IDM_notations}
\end{table}

\section{Intersection Driving Simulation}\label{sec:simulation}
We introduce an Intelligent Intersection Driver Model for simulating low-level vehicle kinematics and pedestrian behaviors that consider the interactions between traffic participants. We then develop a simulator of a partially controlled intersection that involves vehicles and pedestrians.

\subsection{Intelligent Intersection Driver Model (IIDM)}

We develop an Intelligent Intersection Driver Model (IIDM) based on the canonical Intelligent Driver Model (IDM) \cite{treiber2000congested}, a one-dimensional car-following model with tunable parameters \cite{moradipari2022} that drives along a reference path. 
In the canonical IDM, the longitudinal position and velocity in Fren\'{e}t coordinates are computed by
\begin{align}
	\label{eqn:idm1}
	\frac{dv}{dt} &= a_\text{max} \left[1 - \left(\frac{v}{v^*}\right)^{\delta} - \left(\frac{s^*(v, \Delta v)}{s}\right)^2\right], \\
	\label{eqn:idm2}
	&s^*(v, \Delta v) = s_0 + Tv + \frac{v \Delta v}{2 \sqrt{a_\text{max} b_\text{comf}}},
\end{align}
where the variables and constants are introduced in Table \ref{tab:IDM_notations}.
Eq. (\ref{eqn:idm1})--(\ref{eqn:idm2}) serve as a low-level vehicle kinematics model. 

To consider other dynamic agents that may be relevant to a certain vehicle, we define three types of interactions: \texttt{Yield}, \texttt{Not Yield}, and \texttt{Follow}.
\texttt{Yield} is defined as slowing down until a complete stop to avoid collisions when conflict exists. This applies to 1) the vehicles whose future paths intersect with a crosswalk with crossing pedestrians; 2) the vehicles that encounter unyielding crossing traffic and need to avoid collisions. This can be implemented by placing a virtual static leading vehicle at the stop line or at the conflict point, and the simulated vehicle moves according to Eq. (\ref{eqn:idm1})--(\ref{eqn:idm2}).
\texttt{Not Yield} is defined as passing the conflict point first without slowing down or stopping when two vehicles have a conflict in their future paths.
\texttt{Follow} is defined to describe a pair of vehicles that move along the same reference path, where Eq. (\ref{eqn:idm1})--(\ref{eqn:idm2}) can be directly applied.

\subsection{Driving Simulator}
We develop a simulator of vehicles and pedestrians in a partially controlled intersection with two-way stop signs as illustrated in Fig. \ref{fig:teaser}.
The ego vehicle is randomly initialized on a branch with stop signs and the crossing traffic is not constrained. 
Multiple simulated vehicles drive in the crossing traffic lanes and the opposing direction, and multiple pedestrians walk on sidewalks and crosswalks. 

For the simulated vehicles, a human driver is sampled to be \textsc{Aggressive} or \textsc{Conservative} uniformly at the beginning of the episode. 
Then, the driver is sampled to have an intention \textsc{Yield} or \textsc{Not Yield} with $P(\textsc{Yield} \mid \textsc{Conservative}) = 0.9$ and $P(\textsc{Yield} \mid \textsc{Aggressive}) = 0.1$. 
We imitate the fact that both aggressive and conservative drivers may choose to yield or not due to the inherent randomness in human decisions regardless of their traits.
\rv{The general rationale in our simulation design is to differentiate the driving styles of human drivers with different internal states (i.e., trait and intention) to mimic real-world traffic with diverse human drivers.
The differences between heterogeneous driver behaviors on the horizontal lanes are as follows, where the desired speed of each vehicle is sampled from a Gaussian distribution:}
\rv{
\begin{itemize}
    \item Aggressive and non-yielding drivers have a desired speed around \SI{9.0}{m/s} and a minimum distance from the leading vehicle of \SI{4.5}{m}--\SI{7.5}{m}.
    \item Aggressive and yielding drivers have a desired speed around \SI{8.8}{m/s} and a minimum distance from the leading vehicle of \SI{4.8}{m}--\SI{7.8}{m}.
    \item Conservative and non-yielding drivers have a desired speed around \SI{8.6}{m/s} and a minimum distance from the leading vehicle of \SI{5.7}{m}--\SI{8.7}{m}. 
    \item Conservative and yielding drivers have a desired speed around \SI{8.4}{m/s} and a minimum distance from the leading vehicle of \SI{6.0}{m}--\SI{9.0}{m}. 
\end{itemize}
}
The other constants in Eq. (\ref{eqn:idm1})--(\ref{eqn:idm2}) are shared by all categories.
Moreover, the aggressive vehicles on the vertical lane in the opposite direction to the ego vehicle will proceed to cross the intersection whenever a conservative horizontal vehicle comes while conservative vehicles on the vertical lane will stay still until the ego vehicle completes the left turn.

We also add simulated pedestrians on the crosswalks and sidewalks. We assume that pedestrians always have the highest right of way and move with constant speed unless another agent is directly in front of the pedestrians, in which case the pedestrians stay still until the path is clear.
Developing and integrating more realistic and interactive pedestrian behavior models is left for future work.
In general, all vehicles should yield to pedestrians whenever there is a conflict between their future paths within a certain horizon.

\section{Problem Formulation}

We formulate the autonomous navigation of the ego vehicle as a POMDP similar to the problem formulation in our prior work \cite{ma2021reinforcement}. 
The POMDP components are defined as follows: 
\begin{itemize}
	\item \textit{State}: Assume that there are $N$ surrounding vehicles in the scene, $\mathbf{x} = \left[\mathbf{x}^0, \mathbf{x}^1, \mathbf{x}^2,\ldots, \mathbf{x}^N \right]$ denotes the physical state where $\mathbf{x}^0 = \left[x^0, y^0, v^0_x, v^0_y, b^0 \right]$ denotes the ego vehicle state including position, velocity and one-hot indicator of agent type (i.e., vehicle/pedestrian), and $\mathbf{x}^i = \left[x^i, y^i, v^i_x, v^i_y, b^i \right], i\in \{1,\ldots, N\}$ denotes the state of the $i$-th surrounding agent.
	The internal state of the surrounding drivers is represented as $\mathbf{z} = \left[\mathbf{z}^1, \mathbf{z}^2,\ldots, \mathbf{z}^N \right]$. The internal state of each human driver includes two components: $\mathbf{z}_1^i \in \{\textsc{Conservative, Aggressive}\}$ and $\mathbf{z}_2^i \in \{\textsc{Yield, Not Yield}\}$. 
	Assume that there are $M$ pedestrians in the scene, then their physical states are denoted as $\mathbf{x}^{N+1:N+M} = \left[\mathbf{x}^{N+1}, \mathbf{x}^{N+2},\ldots, \mathbf{x}^{N+M} \right]$.
	The joint state is represented by
	\begin{align*}
		\mathbf{s} = \left[\mathbf{x}^0, (\mathbf{x}^1,\mathbf{z}^1),\ldots, (\mathbf{x}^N,\mathbf{z}^N), \mathbf{x}^{N+1},\ldots, \mathbf{x}^{N+M} \right].
	\end{align*}
	
	\item \textit{Observation}: The physical states of all the surrounding vehicles and pedestrians are observable to the ego vehicle while the internal states are not. Formally, the observation is represented by $\mathbf{o} = [\hat{\mathbf{x}}^0, \hat{\mathbf{x}}^1,\ldots, \hat{\mathbf{x}}^{N+M}]$, where $\hat{\mathbf{x}}^i$ is obtained by adding a noise sampled from a zero-mean Gaussian distribution \rv{with a standard deviation of 0.05} to the actual position and velocity to simulate sensor noise.

	\item \textit{Action}: \rv{The action $a \in \{0.0, 1.0, 4.5\} \SI{}{m/s}$ is defined as the target velocity of the ego vehicle for the low-level PD controller to track during the turning process.}
	
	\item \textit{Transition}: The interval between consecutive simulation steps is \SI{0.1}{s}. The behaviors of surrounding vehicles and pedestrians are introduced in Section \ref{sec:simulation}.
	We control the vehicle with a longitudinal PD controller in the same way as our prior work \cite{ma2021reinforcement}, following the left-turn reference path and tracking the target speed determined by the ego policy. We also apply a safety check to make an emergency brake if the distance between the ego vehicle and other agents is too small.
	The episode ends once the ego vehicle completes the left turn successfully, a collision happens, or the maximum horizon is reached.
	
	\item \textit{Reward}: We design a reward function that encourages the driving policy to control the ego vehicle to turn left safely at the intersection as fast as possible without collisions. Formally, the reward function is written as
	\begin{align}
		R(s, a) = \ind\{s \in S_{\text{goal}}\}r_{\text{goal}} +  \ind\{s \in S_{\text{col}}\}r_{\text{col}} + r_{\text{speed}}(s),\nonumber
	\end{align}
	where $r_{\text{goal}} = 2$ and $S_{\text{goal}}$ is a set of goal states where the ego vehicle completes a left turn successfully; $r_{\text{col}} = -2$ and $S_{\text{col}}$ is a set of failure states where a collision happens; and $r_{\text{speed}}(s) = 0.01 \frac{\|v_{\text{ego}}\|}{\SI{4.5}{m/s}}$ is a small reward on the ego vehicle's speed to encourage efficient driving.
\end{itemize}

\section{Deep Reinforcement Learning with\\ Internal State Inference}
In this section, we introduce five variants of deep reinforcement learning with different configurations of human internal state inference for autonomous navigation in complex interactive scenarios (see Fig. \ref{fig:framework_variants}).
The major differences between these architectures lie in the training strategies and the way to incorporate the internal state inference network into the base DRL framework.

\subsection{Internal State Inference}

Consider an urban traffic scenario with the presence of the ego vehicle, $N$ surrounding vehicles, and $M$ pedestrians, where the ego vehicle is controlled by the reinforcement learning policy and the surrounding vehicles are controlled by $N$ human drivers defined in the simulator. 
Let $\mathbf{x}_t$ denote the physical state of all the vehicles and pedestrians at time step $t$, we model the action distribution of the $i$th human driver as $p \left(\mathbf{a}^i_t \mid \mathbf{x}_t, \mathbf{z}^i_t \right)$, where $\mathbf{z}^i_t$ represents the driver's internal state: trait (i.e., aggressive/conservative) and intention (i.e., yield/not yield to the ego vehicle).

Inferring the internal state of the surrounding drivers leads to several advantages. 
First, a discrete internal state is efficient to learn and simple to be integrated into the control policy.
Second, in many situations, the internal state provides even more distinguishable information than predicting their future trajectories. 
For example, in the intersection scenario shown in Fig. \ref{fig:teaser}, the predicted trajectories of the conservative and aggressive vehicles could be similar at the moment before the ego vehicle approaches the intersection, which may not be able to indicate their driving traits effectively. 
However, their traits can be inferred by observing their interaction histories with other vehicles. 
In such cases, the internal state provides the key information explicitly for the ego decision making.

The goal of internal state inference is to determine the distribution $p \left(\mathbf{z}^i_t \mid \mathbf{o}_{1:t} \right)$, where $\mathbf{o}_{1:t}$ denotes the ego agent's historical observations up to time $t$. 
We assume that the ground truth internal states of the surrounding human drivers are available from the simulator at training time and unknown at testing time. 
Thus, the internal state inference module (i.e., a neural network) can be trained by standard supervised learning as a classification task.
By using the information provided by the internal state labels, the auxiliary trait and intention inference tasks provide additional supervision signals in addition to a standard reinforcement learning framework.

\subsection{Graph-Based Representation Learning}\label{sec:encoder}

\begin{figure}[!tbp]
	\centering
	\includegraphics[width=\columnwidth]{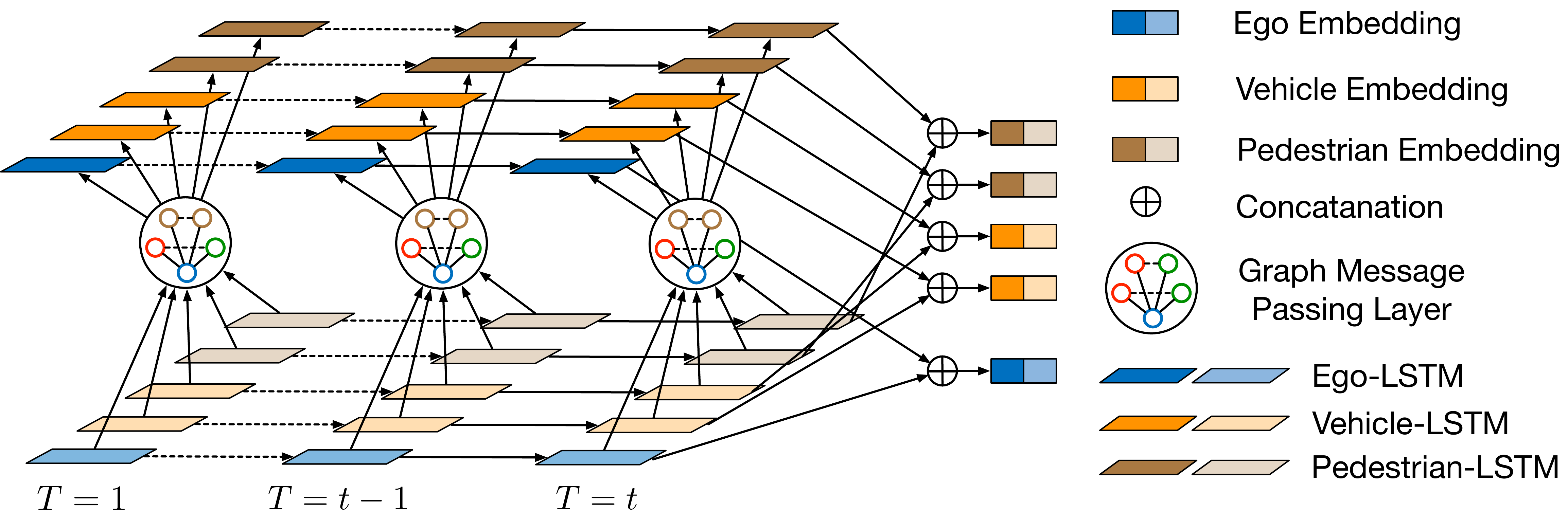}
	\caption{A general diagram of the graph-based encoder, which consists of two LSTM layers and a graph message passing layer in between. We use different LSTM networks to extract features for the ego vehicle, surrounding vehicles, and pedestrians, respectively. The outputs of the two LSTM layers are concatenated to generate their final embeddings. Best viewed in color.}
	\label{fig:encoder}
\end{figure}

A human driver's behavior in complex and dense traffic scenarios is heavily influenced by its relations to other traffic participants. 
The dependence between traffic participants can be naturally formalized as a graph where the nodes represent agents and the edges represent their relations or interactions. 
In a four-way intersection scenario, each vehicle can be potentially influenced by any surrounding agents.
Based on this intuition, we represent the intersection scenario at time $t$ as a fully connected graph $\mathcal{G}_t = (\mathcal{V}_t,\mathcal{E}_t)$, where the node set $\mathcal{V}_t$ contains the nodes for all the vehicles and pedestrians in the scene, and the edge set $\mathcal{E}_t$ contains all the directed edges between each pair of agents. The edges are designed to be directed because the influence between a pair of agents may not be symmetric. Bidirectional relations should be modeled individually.
For example, the leading vehicle tends to have a strong influence on the behavior of the following ones in the same lane. However, the following vehicles merely have an influence on the leading one. The asymmetry also applies to situations where two conflicting agents have different priorities of the right of way such as vehicle-pedestrian interactions.

We adopt a three-layer network architecture similar to the encoder of STGAT~\cite{huang2019stgat} to process both the spatial relational information in $\mathcal{G}_t$ with the graph message passing layer and the temporal information in $o_{1:t}$ with the LSTM recurrent network layer, which is shown in Fig. \ref{fig:encoder}. 
At time step $t$, the observation on the $i$-th vehicle $\mathbf{o}^i_t$ and its observation history $\mathbf{o}^i_{1:t-1}$ are fed into the bottom-level Vehicle-LSTM with a hidden state $\mathbf{h}^i_t$. 
The Vehicle-LSTM parameters are shared among all the vehicles except the ego vehicle.
Similarly, we use another shared Pedestrian-LSTM to extract historical features for pedestrians.
We have
\begin{align}
	\mathbf{v}^0_t =& \ \text{Ego-LSTM}^1 (\mathbf{o}^0_t; \mathbf{h}^0_t), \nonumber\\
	\mathbf{v}^i_t =& \ \text{Vehicle-LSTM}^1 (\mathbf{o}^i_t; \mathbf{h}^i_t), \  i \in \{1,\dots,N\}, \\
	\mathbf{v}^i_t =& \ \text{Pedestrian-LSTM}^1 (\mathbf{o}^i_t; \mathbf{h}^i_t), \ i \in \{N+1,\dots,N+M\}, \nonumber
\end{align}
where $\mathbf{v}^0_t$ and $\mathbf{v}^i_t$ denote the extracted feature vectors of ego and surrounding traffic participants, which encode their historical behaviors. $\text{Ego-LSTM}^1$, $\text{Vehicle-LSTM}^1$, and $\text{Pedestrian-LSTM}^1$ denote the LSTM units at the bottom layer.

The extracted features $\mathbf{v}^0_t$ and $\mathbf{v}^i_t$ are used as the initial node attributes of the corresponding agents in $\mathcal{G}_t$.
We explore the effectiveness of three typical graph message passing layers to process the information across the graph: GAT \cite{velivckovic2018graph}, GCN \cite{kipf2016semi}, and GraphSAGE \cite{hamilton2017inductive}. \rv{The detailed operations of different message passing layers are introduced in Section \ref{sec:appendix}.
The message passing procedures can be applied multiple times to aggregate information from more distant nodes in the graph. Based on the experimental results, we select GAT for message passing, which is written as}
\begin{align}
    \alpha^{ij}_t = \frac{\exp ( \text{LeakyReLU} ( \mathbf{a}^\top [ \mathbf{W}\mathbf{v}^i_t \| \mathbf{W}\mathbf{v}^j_t ] ) )}{\sum_{k \in \mathcal{N}^i} \exp ( \text{LeakyReLU} ( \mathbf{a}^\top [ \mathbf{W}\mathbf{v}^i_t \| \mathbf{W}\mathbf{v}^k_t ] ) )},
\end{align}
where $\mathbf{a}$ and $\mathbf{W}$ denote a learnable weight vector and a learnable weight matrix, $\mathcal{N}^i$ denotes the direct neighbors of node $i$. The symbols $\cdot^\top$ and $\|$ denote transposition and concatenation operations, respectively.
The updated node attributes can be obtained by 
\begin{align}
    \bar{\mathbf{v}}^i_t = \sigma \left(\sum_{j \in \mathcal{N}^i} \alpha^{ij}_t \mathbf{W} \mathbf{v}^j_t \right),
\end{align}
where $\sigma(\cdot)$ denotes a nonlinear activation function. 
	
The updated node attributes are then fed into the top-level LSTM networks with the same parameter-sharing strategy, which is written as
\begin{align}
	\tilde{\mathbf{v}}^0_t =& \ \text{Ego-LSTM}^2 (\bar{\mathbf{v}}^0_t; \bar{\mathbf{h}}^0_t), \nonumber\\
	\tilde{\mathbf{v}}^i_t =& \ \text{Vehicle-LSTM}^2 (\bar{\mathbf{v}}^i_t; \bar{\mathbf{h}}^i_t), \ i \in \{1,\dots,N\},\\
	\tilde{\mathbf{v}}^i_t =& \ \text{Pedestrian-LSTM}^2 (\bar{\mathbf{v}}^i_t; \bar{\mathbf{h}}^i_t), \ i \in \{N+1,\dots,N+M\},\nonumber
\end{align}
where $\text{Ego-LSTM}^2$, $\text{Vehicle-LSTM}^2$, and $\text{Pedestrian-LSTM}^2$ denote the LSTM units at the top layer. The variables $\bar{\mathbf{h}}^0_t$ and $\bar{\mathbf{h}}^i_t$ are hidden states.
The final feature embedding of agent $i$ at time $t$ is obtained by a concatenation of $\mathbf{v}^i_t$ and $\tilde{\mathbf{v}}^i_t$, which encodes both the self-attribute and social-attribute. 

Finally, a multi-layer perceptron (MLP) takes the final embeddings of surrounding vehicles as input and outputs the probability of the corresponding human driver's traits (i.e., aggressive/conservative) and intentions (i.e., yield/not yield).
Note that the pedestrian node attributes are used for message passing yet they are not used for internal state inference. Modeling the internal state of pedestrians is left as future work.

\begin{figure}[!tbp]
	\centering
	\includegraphics[width=0.97\columnwidth]{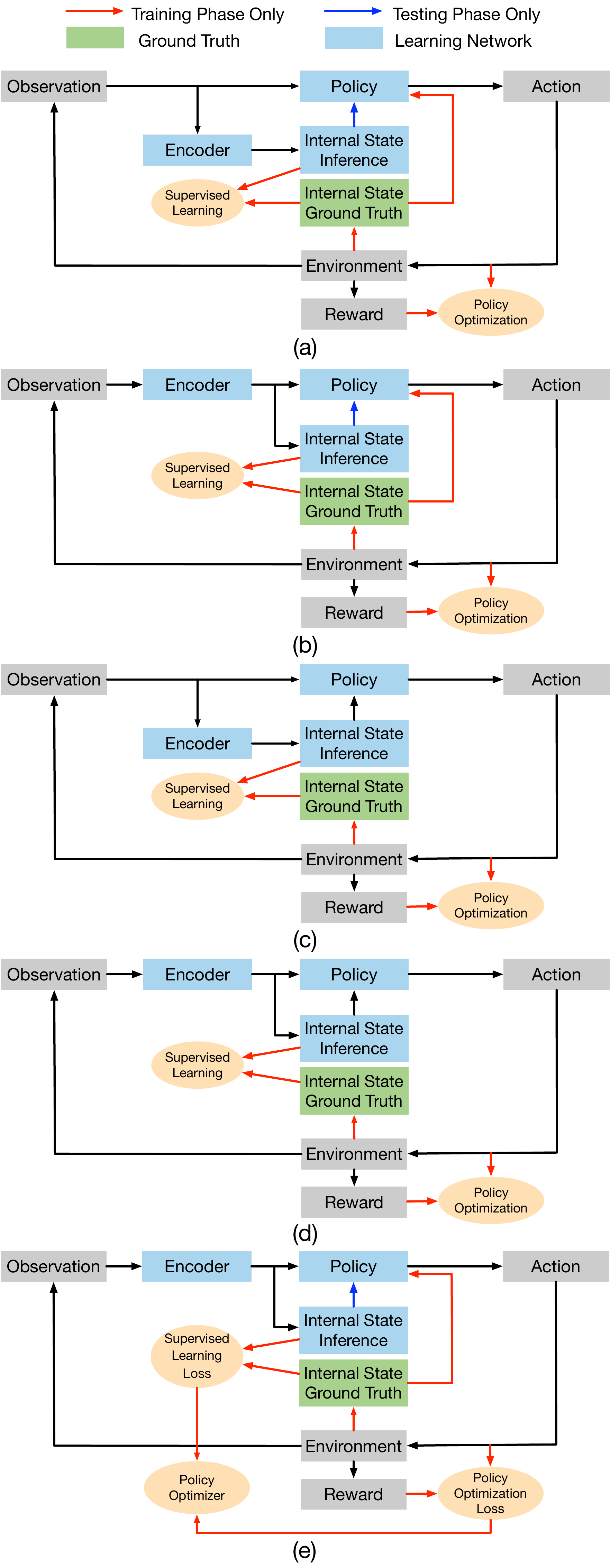}
    \vspace{-0.3cm}
	\caption{Variants of deep reinforcement learning framework architectures with human internal state inference. The gray boxes denote the components in a standard reinforcement learning framework. The blue boxes denote the learnable neural network modules. The green boxes denote the ground truth information that is only accessible during training time. The yellow ellipses denote learning algorithms. \rv{The ground truth information is obtained from the simulator environment during training.}}
\label{fig:framework_variants}
\end{figure}

\subsection{Framework Architectures}

We propose to integrate the human internal state inference into the standard RL-based autonomous navigation framework as an auxiliary task.
The integration can be done in multiple ways.
In this work, we investigate five variants of framework architectures, which are shown in Fig. \ref{fig:framework_variants}.
In all the variants, ground truth internal states can be obtained from the environment (i.e., driving simulator) during training, and the internal state inference network is trained with cross-entropy loss. 
Specific details and the differences among these variants are elaborated below.
Through the comparison between these variants, we can figure out the best combination of model integration and training strategies.
\rv{
\begin{itemize}
	\item \textit{Configuration (a)}: The policy network and the internal state inference network are treated as two separate modules without mutual influence during training.
	During training, the policy network takes in historical observations and true internal states that provide the actual traits and intentions of surrounding vehicles. The graph-based encoder is only used for internal state inference. The policy is refined by a policy optimization algorithm. Meanwhile, the internal state inference network is trained by standard supervised learning separately. During testing, the policy network takes in the inferred internal states. This variant decouples policy learning and internal state inference, which requires individual optimization.
	\item \textit{Configuration (b)}: The difference from configuration (a) is that the policy network and internal state inference network share the same encoder in both training and testing. Thus, the two networks can influence each other via the shared encoder. Both supervised learning loss and policy optimization loss can update the encoder.
	\item \textit{Configuration (c)}: The difference from configuration (a) is that the policy network uses the inferred internal states in both training and testing. The internal state labels are only used to train the inference network. The quality of the information about internal states the policy network uses highly depends on the inference accuracy.
	\item \textit{Configuration (d)}: The difference from configuration (c) is that the policy network and internal state inference network share the same encoder in training and testing.
	\item \textit{Configuration (e)}: The difference from configuration (b) is that the losses from two tasks are coupled by a weighted sum and all the networks are trained with the policy optimizer. This variant enables the highest correlation between the policy network and the internal state inference network.
\end{itemize}
}

\textit{Configuration (a)} contains the following procedures.
The internal state inference network learns the mapping from the historical observations to a latent distribution, i.e. $p_\psi(\mathbf{z}^i_t \mid \mathbf{o}_{1:t})$ where $\psi$ denotes the parameters of the inference network that is trained to minimize the negative log-likelihood:
\begin{align}\label{eqn:isi_obj}
	L(\psi)=-\mathbb{E}_{\mathbf{z}^i_t, \mathbf{o}_{1:t} \sim D} \left[\log p_\psi \left(\mathbf{z}^i_t \mid \mathbf{o}_{1:t}\right)\right],
\end{align}
where the latent state $\mathbf{z}^i_t$ and the historical observations $\mathbf{o}_{1:t}$ are randomly sampled from a replay buffer containing exploration experiences.
The policy takes both the historical observations and the internal state as inputs, i.e. $\pi_\theta \left(a \mid \mathbf{o}_{1:t}, \mathbf{z}^{1:N}_t\right)$ where $\theta$ denotes the policy network parameters trained by the augmented policy optimization objective:
\begin{align}
	\label{eqn:ppo2}
	L(\theta)=\hat{\mathbb{E}}[\min(&\frac{\pi_\theta \left(a \mid  \mathbf{o}_{1:t},\mathbf{z}^{1:N}_t \right)}{\pi_{\theta'} \left(a \mid \mathbf{o}_{1:t}, \mathbf{z}^{1:N}_t \right)}\hat{A}, \nonumber\\
	&\text{clip}(\frac{\pi_\theta(a \mid \mathbf{o}_{1:t},\mathbf{z}^{1:N}_t)}{\pi_{\theta'}(a \mid \mathbf{o}_{1:t},\mathbf{z}^{1:N}_t)},1-\epsilon,1+\epsilon)\hat{A})].
\end{align}

Based on the experimental results in Section \ref{sec:experiments}, we conclude that \textit{Configuration (a)} performs the best among the five variants, which has the following benefits.
First, feeding the ground truth internal state at exploration (i.e., training) time helps the control policy find the trajectory leading to the task goal. This is especially important when the task is difficult and the reward is sparse. 
Second, by using a separate network for each task, the mutual influence of the gradients from different tasks can be minimized. Such mutual influence could be harmful as shown in our experiments. 
Third, by modularizing the two learning modules, our framework allows for flexible choices of network structures in different modules. 

\section{Incorporating Trajectory Prediction and Interactivity Estimation}

\begin{figure}[!tbp]
	\centering
	\includegraphics[width=0.95\columnwidth]{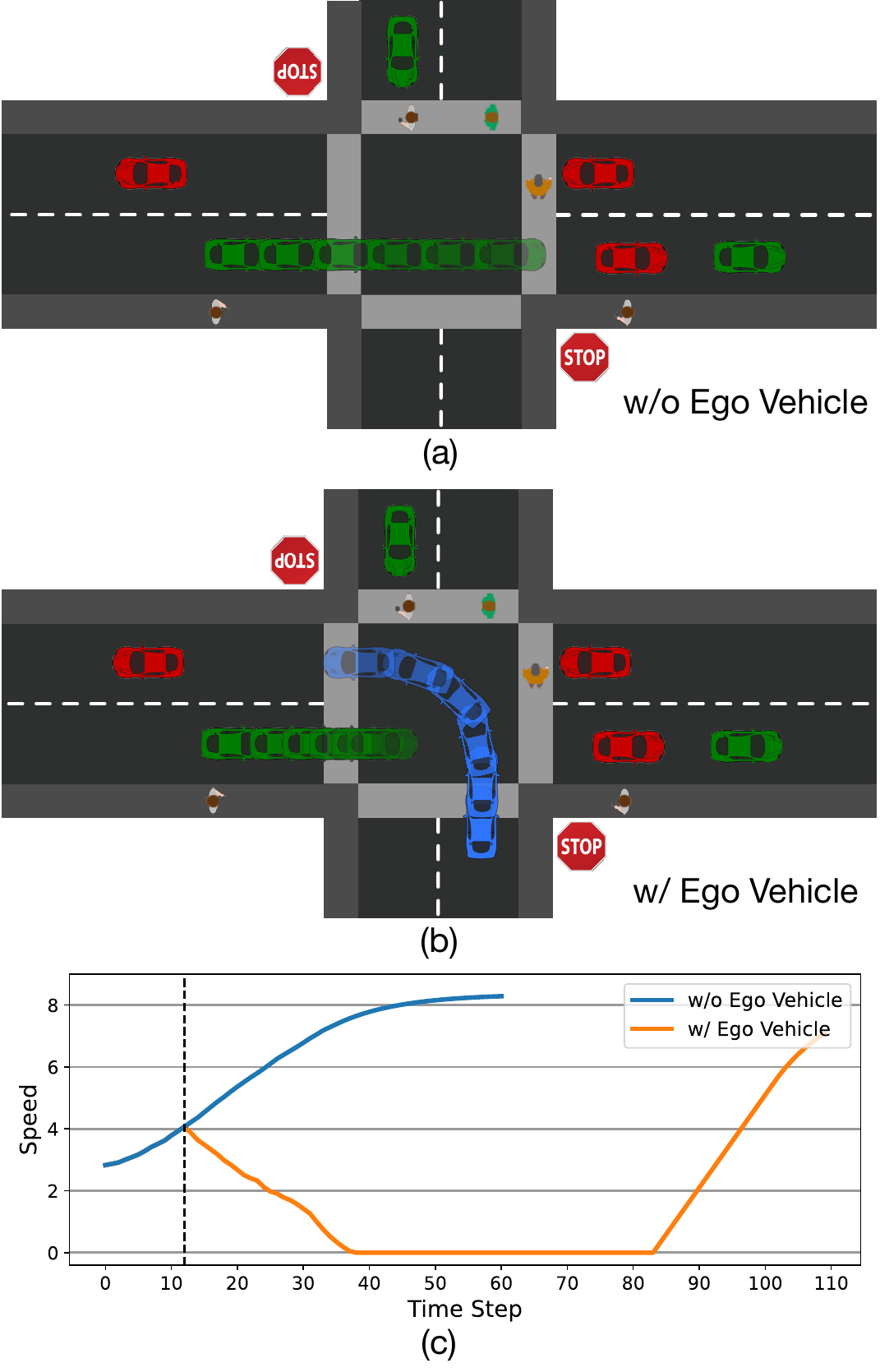}
	\caption{\rv{An illustrative example where the surrounding green vehicle has very different future behaviors in the scenarios without and with the existence of the ego (blue) vehicle. (a) Without the ego vehicle. (b) With the ego vehicle. A comparison of the speed curves of the affected green vehicle in the two settings is shown in (c), which indicates the difference in behaviors between the two settings. The future trajectories of other traffic participants are omitted.}}
	\label{fig:ego_diff}
\end{figure}

Besides inferring the high-level internal states of human drivers, the autonomous navigation task also benefits from forecasting their future trajectories that provide fine-grained behavioral cues as well as reasoning about the potential influence of the ego vehicle on surrounding agents. 
In this work, we design an auxiliary trajectory prediction task to infer how the other traffic participants will behave in the presence of the ego vehicle.
Moreover, in complex urban traffic, human drivers tend to implicitly estimate to what extent they could influence the behaviors of other traffic participants to enhance situational awareness and facilitate their negotiation and driving efficiency. Motivated by this intuition, we design a mechanism to estimate the interactivity scores of other agents that can be used by the policy network.

\rv{Here we provide an example to illustrate the core concept.
In Fig. \ref{fig:ego_diff}(a), the green vehicle on the left side can speed up to cross the intersection without any conflict.
However, in Fig. \ref{fig:ego_diff}(b), the green vehicle must slow down and yield to the blue ego vehicle to avoid a collision.
The difference in the green vehicle's behavior in these two situations can be used to compute the influence of the ego vehicle.
Meanwhile, the speed profiles of the same green vehicle in the two settings are compared in Fig. \ref{fig:ego_diff}(c) for a quantitative illustration.}
We propose to predict the future trajectories of other agents in both situations and quantify the difference as interactivity scores, which are used as input to the ego policy network.
Moreover, since the ego vehicle tends to negotiate with the agents with large interactivity scores, the trajectory prediction of those agents needs to be more accurate than those with small interactivity scores to ensure better safety and efficiency.
Therefore, we propose a weighting strategy in the prediction loss based on the interactivity scores to encourage better prediction of important agents that may have strong interactions with the ego vehicle.
In both training and testing scenarios, the ego vehicle always exists and may influence the other agents, thus the prediction in the situation in Fig. \ref{fig:ego_diff}(a) (i.e., without the ego vehicle) can be treated as counterfactual reasoning.

\subsection{Trajectory Prediction}\label{sec:traj_pred}
The trajectory prediction task is formulated as a regression problem solved by supervised learning, where the ground truth future trajectories can be obtained by simulation. This can provide additional supervision signals to refine the graph representation learning in the encoder and thus help with the improvement of other downstream components.
We forecast the future trajectories of surrounding agents in both situations (i.e., without and with the existence of the ego vehicle) through two separate prediction heads.
The former task encourages the model to capture the natural behaviors of surrounding agents defined by the simulation without the intervention of the learned ego vehicle's policy.
The latter task encourages the model to capture how the surrounding agents will react to the ego vehicle's future behavior through their future trajectories.

Formally, we denote the prediction horizon as $T_\text{f}$ and the objective of prediction is to estimate two conditional distributions $p\left(\mathbf{x}^{1:N+M}_{t+1:t+T_\text{f}} \mid \mathbf{o}^{1:N+M}_{1:t}\right)$ (without the ego vehicle) and $p\left(\mathbf{x}^{1:N+M}_{t+1:t+T_\text{f}} \mid \mathbf{o}_{1:t}\right)$ (with the ego vehicle). 
\rv{Without loss of generality, the distributions are assumed to be Gaussian with a fixed diagonal covariance matrix $\bm{\Sigma}$ for simplicity; thus, the model can focus on predicting the mean of distributions. In future work, we will further investigate more complex traffic scenarios where the predicted distributions can be multi-modal (e.g., Gaussian Mixture Model) with learnable covariance.}
\begin{itemize}
    \item To predict future trajectories in the scenarios without the ego vehicle, we propose to pre-train another prediction model branch including a graph-based encoder with the same architecture as the one shown in Fig. \ref{fig:encoder} except that there is no ego vehicle involved as well as an MLP prediction head. The parameters of these networks are fixed without further updates during the formal training stage to generate counterfactual future trajectories. The reason for using a separate prediction branch is to minimize the influence of the ego vehicle in counterfactual prediction.
    \item To predict future trajectories in the scenarios with the ego vehicle, an MLP prediction head takes the final node attributes $\tilde{\mathbf{v}}^i_t$ as input and outputs the means of predicted trajectory distributions of agent $i$ (i.e., $\hat{\bm{\mu}}^{i, \text{w/ Ego}}_{t+1:t+T_\text{f}}$). We use the pre-trained network parameters in the former setting to allow for better initialization.
\end{itemize}

\subsection{Interactivity Estimation}\label{sec:interactivity}
We propose an interactivity estimation mechanism based on the difference between the predicted trajectories in the two situations discussed in Section \ref{sec:traj_pred}.
\rv{The underlying intuition is that the ego vehicle can potentially influence the behavior of surrounding agents that have conflicts in their future paths and negotiate the right of way. The estimated strength of influence indicated by the difference between their future trajectories can quantitatively imply to what extent the ego vehicle can try to interact or negotiate with a certain agent, which is named as an \textit{interactivity score (IS)} and helps the ego vehicle to select a proper occasion to proceed.}

\begin{figure}[!tbp]
	\centering
	\includegraphics[width=\columnwidth]{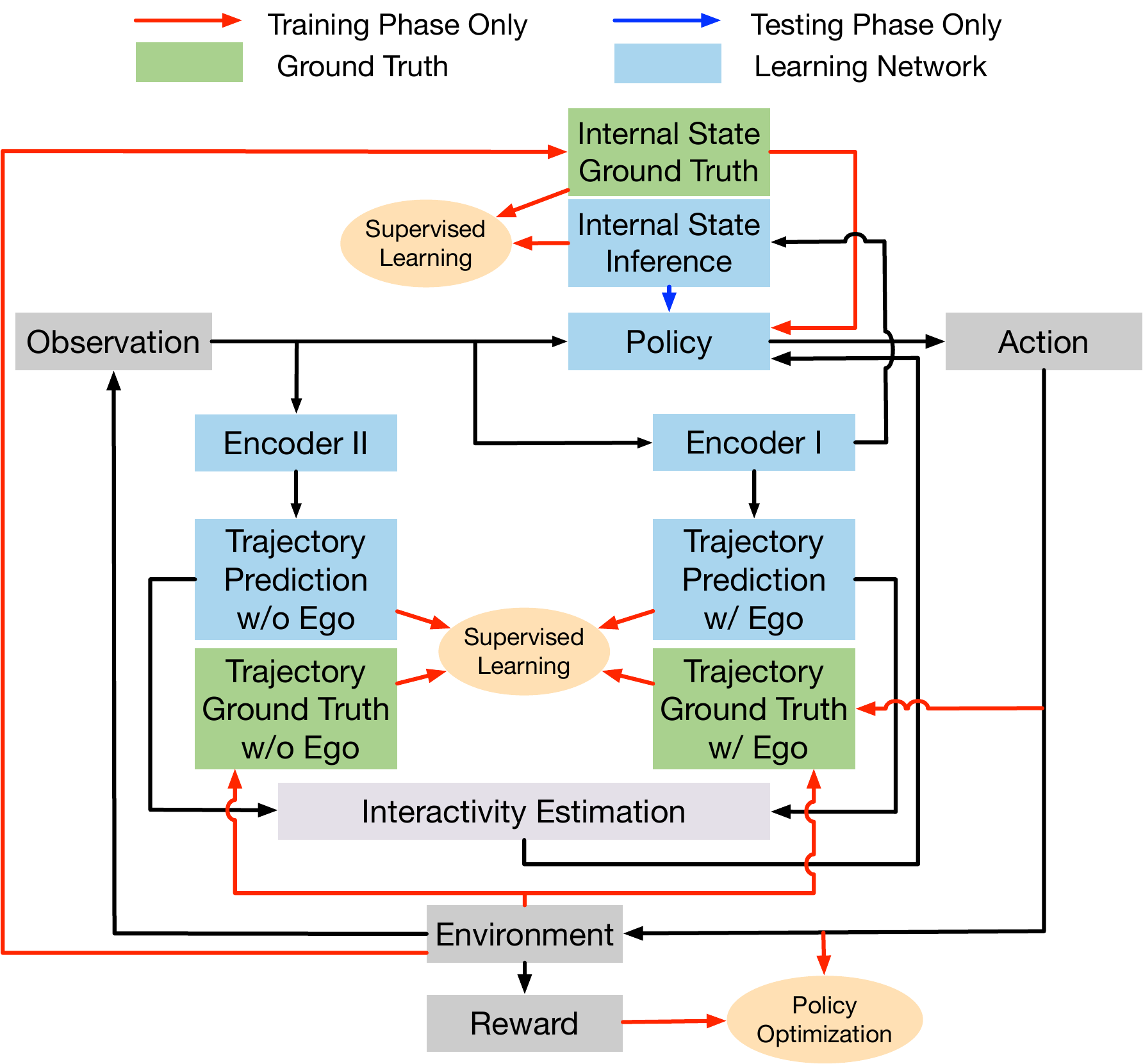}
	\caption{A diagram of the proposed reinforcement learning framework with auxiliary supervised learning tasks. Note that Encoder II and the trajectory prediction head w/o Ego are pre-trained and their parameters are fixed without further update during the formal training stage.}
	\vspace{-0.3cm}
	\label{fig:framework}
\end{figure}

\rv{To indicate the differences in the surrounding agents' future behaviors under the probabilistic setting, we propose to use the Kullback–Leibler (KL) divergence between the two trajectory distributions given by}
\begin{align}
    \mathbf{x}^{i}_{t+1:t+T_\text{f}} \mid \mathbf{o}_{1:t} &\sim \mathcal{N}\left(\hat{\bm{\mu}}^{i, \text{w/ Ego}}_{t+1:t+T_\text{f}}, \bm{\Sigma}\right) = \mathcal{N}\left(\hat{\bm{\mu}}^{i}_1, \bm{\Sigma}\right), \nonumber \\
    \mathbf{x}^{i}_{t+1:t+T_\text{f}} \mid \mathbf{o}^{1:N+M}_{1:t} &\sim \mathcal{N}\left(\hat{\bm{\mu}}^{i, \text{w/o Ego}}_{t+1:t+T_\text{f}}, \bm{\Sigma}\right) = \mathcal{N}\left(\hat{\bm{\mu}}^{i}_2, \bm{\Sigma}\right), \nonumber
\end{align}
to indicate the difference quantitatively, which is computed by
{\small
\begin{align}
    & \ D_{\text{KL}} \left(p\left(\mathbf{x}^{i}_{t+1:t+T_\text{f}} \mid \mathbf{o}_{1:t}\right) \| \ p\left(\mathbf{x}^{i}_{t+1:t+T_\text{f}} \mid \mathbf{o}^{1:N+M}_{1:t}\right)\right) \nonumber \\
    &= \ \frac{1}{2}\left( \text{Tr} \left(\bm{\Sigma}^{-1}\bm{\Sigma}\right) - d + \left(\hat{\bm{\mu}}^{i}_1 - \hat{\bm{\mu}}^{i}_2\right)^\top \bm{\Sigma}^{-1} \left(\hat{\bm{\mu}}^{i}_1 - \hat{\bm{\mu}}^{i}_2\right) + \ln \left( \frac{\det \bm{\Sigma}}{\det \bm{\Sigma}}\right)\right) \nonumber\\
    &= \ \frac{1}{2} \left(\hat{\bm{\mu}}^{i}_1 - \hat{\bm{\mu}}^{i}_2\right)^\top \bm{\Sigma}^{-1} \left(\hat{\bm{\mu}}^{i}_1 - \hat{\bm{\mu}}^{i}_2\right) 
    = \ \frac{1}{2\sigma^2} \| \hat{\bm{\mu}}^{i}_1 - \hat{\bm{\mu}}^{i}_2 \|^2,
\end{align}
}%
where $\sigma^2$ is the constant covariance value in the diagonal of $\bm{\Sigma}$, $d$ is the dimension of the distributions, $\text{Tr}(\cdot)$ denotes the trace of a matrix, $\cdot^\top$ denotes the transpose of a vector, $\bm{\Sigma}^{-1}$ and $\det \bm{\Sigma}$ denote the inverse and determinant of the covariance matrix, respectively.
Due to the Gaussian assumption with fixed covariance, the KL divergence reduces to the $L_2$ distance between the mean vectors of two trajectory distributions multiplied by a constant.
For simplicity, we define the interactivity score $\mathbf{w}^i_t$ of agent $i$ at time $t$ as
\begin{align}\label{equ:IS}
	\mathbf{w}^{i}_t = \left\|\hat{\bm{\mu}}^{i, \text{w/ Ego}}_{t+1:t+T_\text{f}} - \hat{\bm{\mu}}^{i, \text{w/o Ego}}_{t+1:t+T_\text{f}}\right\|^2.
\end{align}

The interactivity scores can be treated as a feature of each agent and used by the policy network.
Moreover, we use them as the weights of prediction errors in the loss function for trajectory prediction, which is computed by
\begin{align}\label{eqn:tp_obj}
    L^\text{TP} = \frac{1}{N+M} \sum_{i=1}^{N+M} \mathbf{w}^{i}_t \cdot \left\| \hat{\bm{\mu}}^{i, \text{w/ Ego}}_{t+1:t+T_\text{f}} - \mathbf{x}^{i, \text{w/ Ego}}_{t+1:t+T_\text{f}} \right\|^2,
\end{align}
where $\mathbf{x}^{i, \text{w/ Ego}}_{t+1:t+T_\text{f}}$ is the ground truth of future trajectories.

\begin{algorithm*}[!tbp]
	\caption{Reinforcement Learning with Auxiliary Tasks (Formal Training Phase)}  
	\begin{algorithmic}[1]  
		\REQUIRE initial parameters of the policy network $\theta_0$,  value function $\phi_0$, clipping threshold $\epsilon$, \\ $\quad$~~ encoder $\psi^{\text{Enc}}_0$, internal state inference network $\psi^{\text{ISI}}_0$, trajectory prediction head considering the ego vehicle $\psi^{\text{TP}}_0$\\
		
		\FOR{$k=1,2,...$}
		\STATE // Collect a set of trajectories $\mathcal{D}_k$ by running the policy $\pi_k = \pi(\theta_k)$ in the environment following the steps below.
		\FOR{$r=1,2,...$}
		\FOR{$t=1,2,...$} 
		\STATE Infer the internal states of surrounding vehicles $\hat{\mathbf{z}}^{1:N}_t$ with the current $\psi^\text{Enc}$ and $\psi^\text{ISI}$.
		\STATE Generate the future trajectory hypotheses $\hat{\bm{\mu}}^{1:N+M, \text{w/ Ego}}_{t+1:t+T_\text{f}}$ with the current $\psi^\text{Enc}$ and $\psi^\text{TP}$.
		\STATE Generate the future trajectory hypotheses $\hat{\bm{\mu}}^{1:N+M, \text{w/o Ego}}_{t+1:t+T_\text{f}}$ with the pre-trained prediction model.
		\STATE Compute interactivity scores of other agents by Eq. (\ref{equ:IS}).
		\STATE Choose an action for ego agent using the policy $\pi_k$ and obtain the next state from the environment.
		\ENDFOR
		\ENDFOR
		\STATE // Update the learnable networks following the steps below.
		\STATE Compute the rewards-to-go $\hat{R}_t$.
		\STATE Compute the advantage estimates $\hat{A}_t$ using any method of advantage estimation with the current value function $V_{\phi_k}$.
		\STATE Update the policy network by maximizing the PPO objective via stochastic gradient ascent with Adam optimizer:
		\begin{align}
			\theta_{k+1} = \arg \max_\theta  \mathbb{E}\left[\min\left(\frac{\pi_\theta \left(a \mid  \mathbf{o}_{1:t},\mathbf{z}^{1:N}_t,\mathbf{w}^{1:N+M}_t \right)}{\pi_{\theta_k} \left(a \mid \mathbf{o}_{1:t}, \mathbf{z}^{1:N}_t,\mathbf{w}^{1:N+M}_t \right)}\hat{A},
			\text{clip}\left(\frac{\pi_\theta(a \mid \mathbf{o}_{1:t},\mathbf{z}^{1:N}_t,\mathbf{w}^{1:N+M}_t)}{\pi_{\theta_k}(a \mid \mathbf{o}_{1:t},\mathbf{z}^{1:N}_t,\mathbf{w}^{1:N+M}_t)},1-\epsilon,1+\epsilon \right)\hat{A}\right)\right].
		\end{align}
		Note that the ground truth internal states $\mathbf{z}^{1:N}_t$ are used as the input of policy network in the training phase while the inferred internal states $\hat{\mathbf{z}}^{1:N}_t$ are only used to compute loss function (\ref{eqn:isi_obj}).
		\STATE Fit the value function by regression on mean squared error via gradient descent:
		\begin{align}
			\phi_{k+1} = \arg \min_\phi \mathbb{E} \left(V_\phi (s_t) - \hat{R}_t \right)^2.
		\end{align}
		\STATE Update the encoder $\psi^{\text{Enc}}$, internal state inference network $\psi^{\text{ISI}}$, and trajectory prediction head $\psi^{\text{TP}}$ by minimizing the corresponding loss function (\ref{eqn:isi_obj}) and (\ref{eqn:tp_obj}) via gradient descent.
		\ENDFOR
	\end{algorithmic}  
	\label{alg:complete_train}
\end{algorithm*}

\begin{algorithm*}[!tbp]
	\caption{Reinforcement Learning with Auxiliary Tasks (Testing Phase)}  
	\begin{algorithmic}[1]  
		\REQUIRE parameters of policy network $\theta$, encoder $\psi^{\text{Enc}}$, internal state inference network $\psi^{\text{ISI}}$, trajectory prediction heads $\psi^{\text{TP}}$\\
		
		\STATE // Given a testing scenario, run the policy following the steps below.
		\FOR{$t=1,2,\ldots$} 
		\STATE Infer the internal states of surrounding vehicles $\hat{\mathbf{z}}^{1:N}_t$ with current $\psi^\text{Enc}$ and $\psi^\text{ISI}$.
		\STATE Generate the trajectory hypotheses $\hat{\bm{\mu}}^{1:N, \text{w/ Ego}}_{t+1:t+T_\text{f}}$ and $\hat{\bm{\mu}}^{1:N, \text{w/o Ego}}_{t+1:t+T_\text{f}}$ with $\psi^\text{Enc}$, $\psi^\text{TP}$ and the pre-trained prediction model.
		\STATE Compute interactivity scores of other agents by Eq. (\ref{equ:IS}).
		\STATE Choose an action $a_t$ for ego agent using the policy $\pi_{\theta}(a_t \mid \mathbf{o}_{1:t},\hat{\mathbf{z}}^{1:N}_t,\mathbf{w}^{1:N+M}_t)$ and obtain the next state.
		\ENDFOR
	\end{algorithmic}  
	\label{alg:complete_test}
\end{algorithm*}

\subsection{Complete Framework}\label{sec:complete_framework}
An overall diagram of the complete method is shown in Fig. \ref{fig:framework}, which integrates auxiliary supervised learning tasks into the reinforcement learning framework.
The detailed pseudocode of the proposed method in the training and testing phases is provided in Algorithm \ref{alg:complete_train} and Algorithm \ref{alg:complete_test}, respectively. 

First, we have a graph-based encoder to extract spatio-temporal features from historical observations of all the agents.
Second, we have an internal state inference module to recognize the traits and intentions of surrounding vehicles.
Third, we have a trajectory prediction module to forecast the future behaviors of other agents with the existence of the ego vehicle. 
The ground truth labels of internal states and future trajectories can be obtained from the environment (i.e., driving simulator) in the training phase, which are not needed in the testing phase.
We also have another pre-trained trajectory prediction module to forecast the future behaviors of other agents without the existence of the ego vehicle.
Fourth, we estimate the interactivity scores of other agents.
Finally, the policy network outputs the action distribution based on the historical observations, inferred internal states, and estimated interactivity scores of surrounding agents.

The explainable aspects of our method come from two auxiliary tasks: (a) internal state inference; and (b) interactivity estimation.
On the one hand, our method can infer the traits and intentions of surrounding vehicles, which can inform the policy network about whether they tend to yield to the ego vehicle. The inferred internal state can serve as an explanation for the decision making.
On the other hand, the estimated interactivity scores can reflect how much influence the ego vehicle can potentially have on surrounding agents. 
A higher interactivity score implies that the ego vehicle has a higher possibility of being able to influence and negotiate with the corresponding agent to improve driving efficiency.

\section{Experiments}\label{sec:experiments}

\subsection{Experiment Settings and Implementation Details}

\rv{We train our method and all the baselines three times with different random seeds} and each trial is trained for $10^7$ environment steps per epoch.
We use a learning rate of $10^{-4}$ for the policy optimizer and $10^{-3}$ for the optimizer of the value baseline and supervised learning.
In the framework variant in Fig. \ref{fig:framework_variants}(e), we set the weight of the supervised learning losses as $0.1$.
\rv{We run all experiments on a Linux workstation with Intel i9-10940X CPU and a NVIDIA Quadro RTX 6000 GPU.}

The encoder consists of six LSTM networks with a hidden size of 64 in the bottom and top layers for different types of agents. The weight vectors/matrices in the graph message passing layers have a proper dimension corresponding to specific feature dimensions.
The internal state inference network and trajectory prediction networks are three-layer MLPs with a hidden size of 64. The policy network is an LSTM network with a hidden size of 64, whose input is a concatenation of the observations on all the agents. \rv{The vehicle and pedestrian features are ordered based on their distances to the ego vehicle (from small to large).}

\subsection{Evaluation Metrics and Baselines}
We evaluate our method with widely used metrics in decision making for autonomous driving (i.e., completion rate, collision rate, timeout rate, time to completion) and internal state inference (i.e., classification accuracy).
We define that an episode is considered to be a successful completion if the ego vehicle completes the left turn within 25 seconds (i.e., 250 time steps) without collision with other traffic participants. Otherwise, the episode is considered a collision or timeout case. Only the completion cases are used to compute the time to completion, which indicates driving efficiency quantitatively.
We compare our full method with state-of-the-art baselines \cite{morton2017simultaneous, ma2021reinforcement, liu2022learning} and several ablation framework settings to demonstrate the effectiveness of each component.
\rv{We evaluate our method and baselines in 1,000 testing scenarios consistently for fair comparisons.}

\subsection{Internal State Inference}

\begin{figure}[!tbp]
	\centering
	\includegraphics[width=\columnwidth]{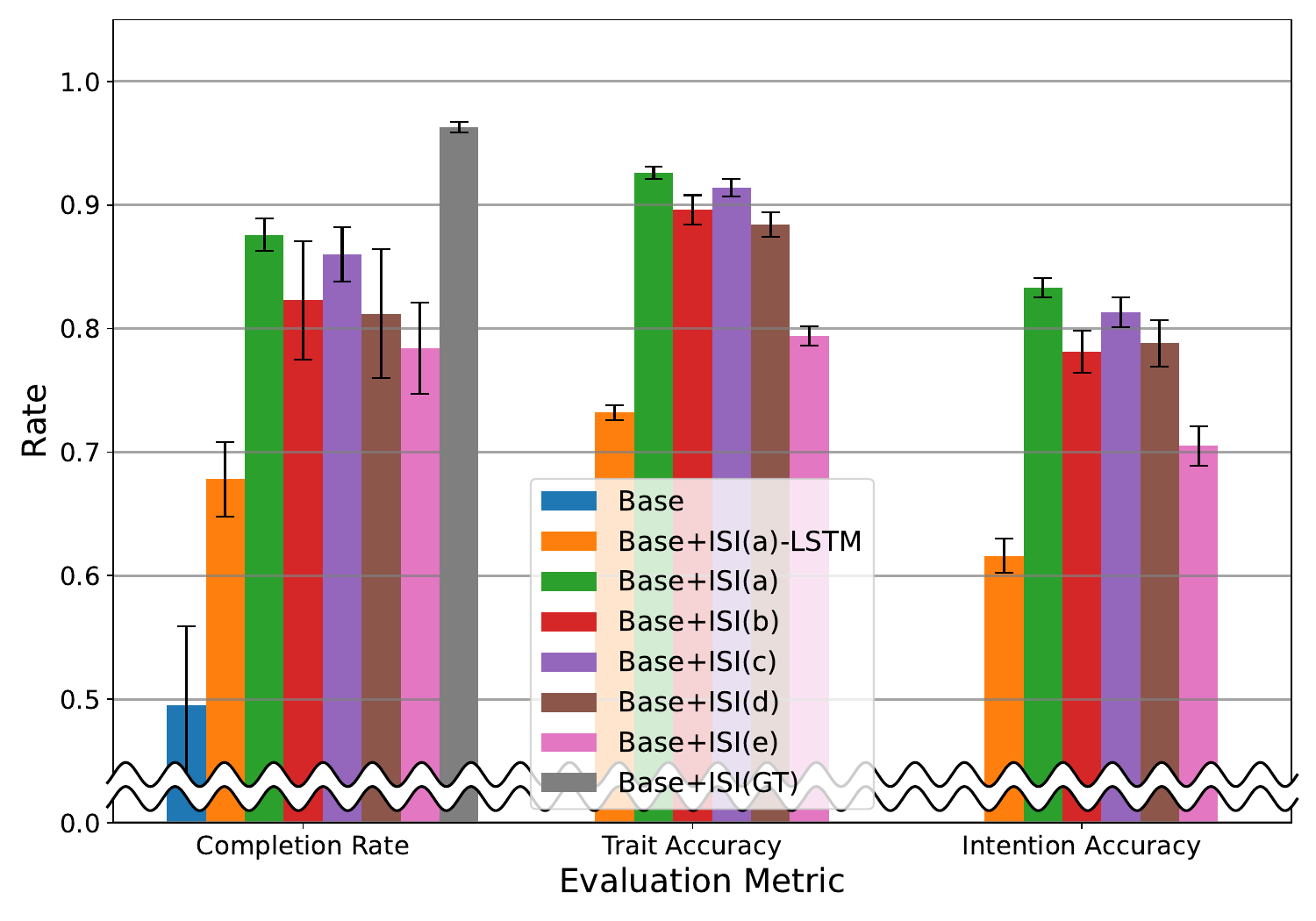}
	\caption{\rv{A comparison of different architectures for integrating internal state inference (ISI) into the Base reinforcement learning framework. The Base method does not have the trait or intention inference. Base+ISI(GT) uses ground truth internal states for both training and testing, which serves as a performance upper bound.} }
 \vspace{-0.2cm}
	\label{fig:ISI_variants}
\end{figure} 

\begin{figure}[!tbp]
	\centering
	\includegraphics[width=\columnwidth]{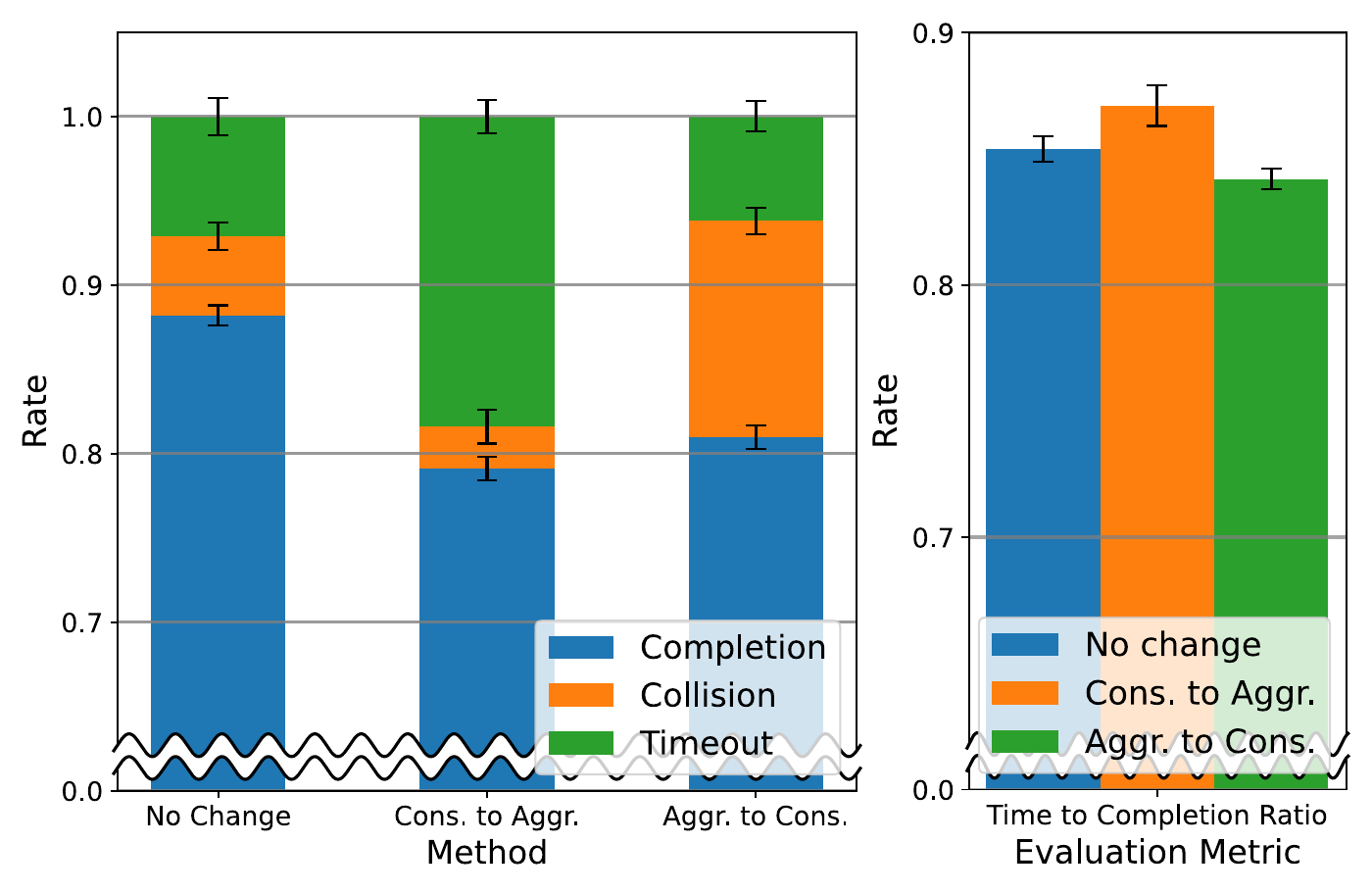}
	\caption{\rv{The comparison of the decision making performance of applying different internal state manipulation operations. ``Cons. to Aggr.'' means randomly changing the inferred internal states of conservative vehicles to aggressive with a probability of $0.5$. ``Aggr. to Cons.'' means the manipulation of inferred internal states in the opposite way.}}
	\label{fig:ISI_manipulation}
\end{figure}

\begin{figure*}[!tbp]
	\centering
	\includegraphics[width=\textwidth]{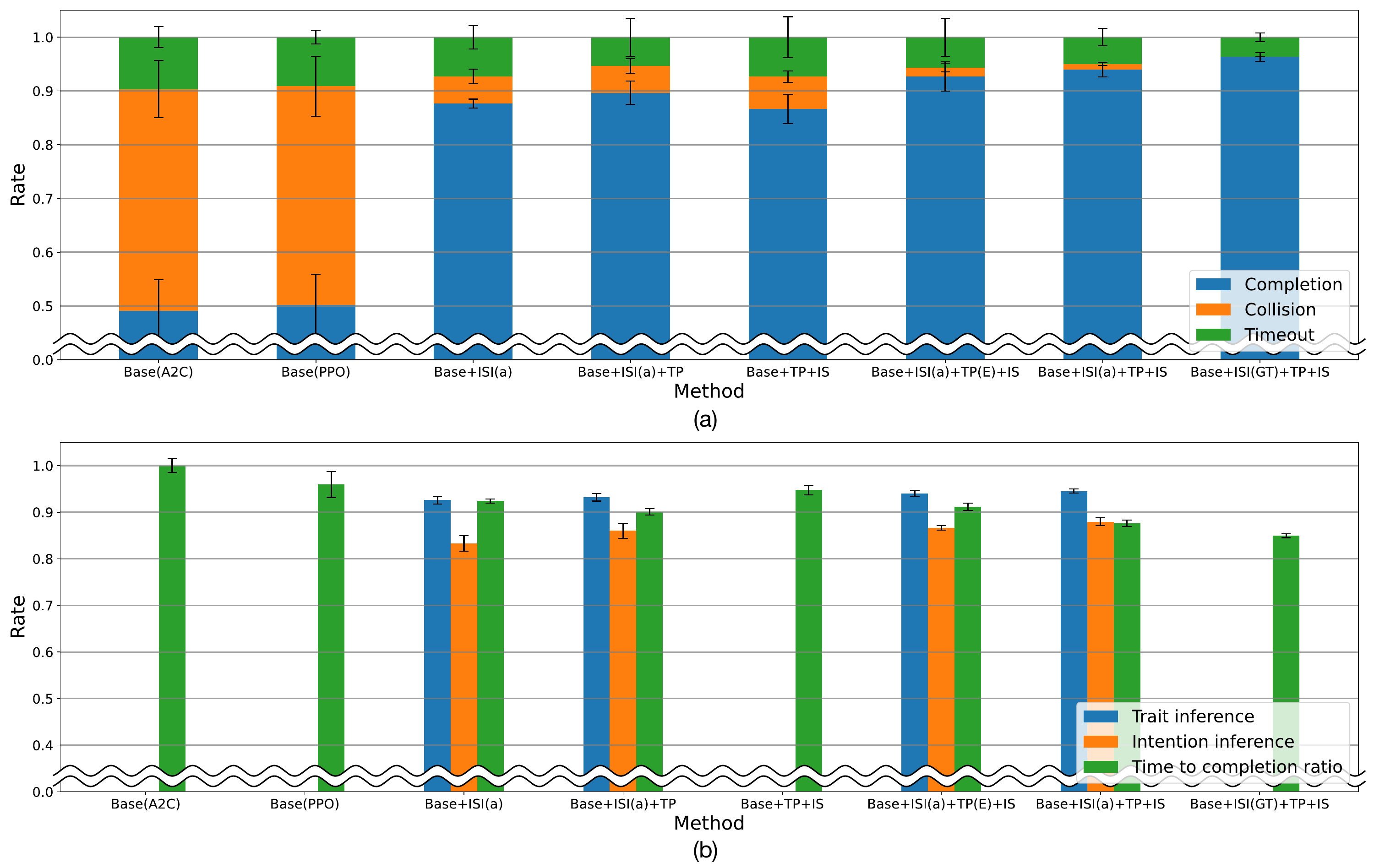}
	\caption{\rv{The comparison of the decision making and internal state inference performance between widely used baseline methods and our ablation settings. Note that TP(E) means that the trajectory predictor is trained with a loss function with equal weights of prediction errors. Base+ISI(a)+TP+IS is our proposed method. Similar to Fig. \ref{fig:ISI_variants}, Base+ISI(GT)+TP+IS uses ground truth internal states for both training and testing, which serves as a performance upper bound. We adopt PPO as the RL algorithm by default except for the Base(A2C) baseline. (a) The comparison of the decision making performance (i.e., completion rate, collision rate, timeout rate) where the three rates sum to one. (b) The comparison of internal state inference performance (i.e., trait accuracy, intention accuracy) and driving efficiency (i.e., time to completion ratio). Note that some methods do not have trait or intention inference modules, thus they are not evaluated on these aspects. Also, we report the time-to-completion ratio with respect to the Base(A2C) baseline instead of the absolute values to indicate driving efficiency. Note that we omit the bar plots in the small number range to better show the differences between various methods.}}
    \vspace{-0.3cm}
	\label{fig:baselines}
\end{figure*}

We demonstrate the effectiveness of internal state inference and compare the performance of the five variants of framework architectures presented in Fig. \ref{fig:framework_variants} through quantitative analysis and ablation study.

The comparison between different variants of framework architectures is shown in Fig. \ref{fig:ISI_variants}. 
\rv{Base+ISI(GT) uses the ground truth internal states as the input of the policy network, which serves as a performance upper bound among all the methods and implies the effectiveness of taking advantage of internal states in the decision making process.}
The Base method is a standard reinforcement learning framework without auxiliary modules.
It shows that the Base method performs the worst among all the variants and different strategies of integrating the ISI module into the base method lead to different degrees of improvement.
\rv{The Base method is only able to capture the behavior patterns of surrounding traffic participants implicitly through the policy network, which makes it difficult to learn the traits and intentions of other vehicles. Thisleads to poor performance with many collisions and lower driving efficiency in our environment.
Base+ISI(a) performs the best among the five variants of architectures, which improves the average completion rate by 76.9\% over the Base method. }

Base+ISI(a)-LSTM simply uses a shared LSTM network instead of a graph-based encoder to extract the historical information of each agent individually. Base+ISI(a) outperforms Base+ISI(a)-LSTM by a large margin in terms of both completion rate and internal state inference, which implies the effectiveness of spatio-temporal behavior modeling of interactive agents.

\rv{The comparison between Base+ISI(a) and Base+ISI(b) implies that having a separate encoder for ISI performs better than using a shared encoder for the policy network and the ISI network in terms of both average completion rate and ISI accuracy. The reason is that the policy network needs to focus more on the ego perspective to make decisions while the ISI network needs to capture explicit relations between interactive agents in a distributed manner. Since a shared encoder needs to learn useful features for both aspects, the encoded information may distract the learning process of both networks. 
We also have a consistent observation by comparing Base+ISI(c) and Base+ISI(d), although they adopt a different strategy in using the internal states during training.}

\rv{The comparison between Base+ISI(a) and Base+ISI(c) implies that using the ground truth internal state labels as the input of the policy network during training performs better than using the inferred internal state in terms of average completion rate while achieving a comparable performance in terms of ISI accuracy. It is reasonable that both architectures achieve a similar ISI accuracy because the learning processes of the ISI network are essentially the same. However, the inferred internal states in Base+ISI(c) may be wrong during training, which misleads policy learning and makes the training process unstable. 
Using the ground truth internal states during training is similar to the teacher forcing technique in \cite{williams1989learning}, which can stabilize training and improve performance.}

\rv{The comparison between Base+ISI(a) and Base+ISI(e) implies that updating the policy network and ISI network separately with individual loss performs better than the coupled training strategy.
The combination of two losses leads to biased gradient estimates for both policy learning and internal state inference tasks. Although Base+ISI(e) outperforms Base in terms of average completion rate, the improvement is relatively minor and cannot reflect the significant benefits of ISI. And the average ISI accuracy is much worse than other variants.}

\rv{By comparing Base+TP+IS and Base+ISI(a)+TP+IS settings in Fig. \ref{fig:baselines}, we can observe that the improvement brought by ISI is also significant even with the auxiliary modules of trajectory prediction and interactivity estimation, which implies the effectiveness of explicit modeling of human drivers' traits and intentions. The reason is that with the inferred internal states of surrounding vehicles, it is easier for the ego driving policy to figure out a proper opportunity to proceed.}

\begin{figure}[!tbp]
	\centering
	\includegraphics[width=\columnwidth]{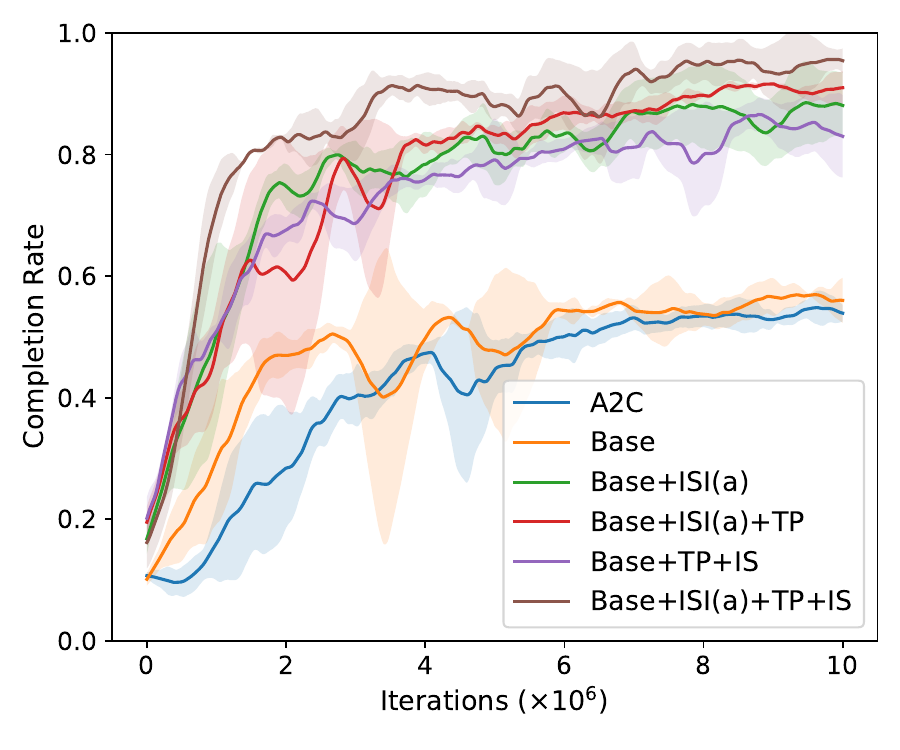}
	\caption{\rv{The comparison of the completion rate curves of different model settings during the training process. The solid lines represent the average completion rates and the shaded areas represent the standard deviations. The comparison between Base and Base+ISI(a) shows the effectiveness of internal state inference, which demonstrates a significant improvement in completion rate. The comparison between Base+ISI(a) and Base+ISI(a)+TP+IS shows the effectiveness of the combination of trajectory prediction and interactivity estimation. Best viewed in color.}}
	\label{fig:learning_curves}
\end{figure}

\rv{The comparison of the completion rate curves of different model settings during the training process is shown in Fig. \ref{fig:learning_curves}.
The results show that the policy learning process tends to be faster and more stable with internal state inference.
The reason is that during the training phase, the RL agent is able to use additional information about the agents in the environment through internal state labels, which helps capture the correlations between observations and the underlying traits and intentions of interactive agents effectively.}

To explicitly demonstrate the influence of internal state inference on the decision making of the ego vehicle, we manipulated the inferred internal states of surrounding vehicles and compared the outcomes in Fig. \ref{fig:ISI_manipulation}. 
It shows that ``Cons. to Aggr.'' leads to a lower collision rate and a higher timeout rate than ``No change'' because, with more inferred aggressive vehicles, the ego vehicle tends to yield and wait until it finds a conservative vehicle to proceed. 
In contrast, ``Aggr. to Cons.'' leads to a higher collision rate and a slightly lower timeout rate than ``No change'' because, with more inferred conservative vehicles, the ego vehicle tends to be more aggressive.
These results imply that the learned policy has a high dependence on the inferred internal states, which performs much better when using the original inference than the manipulated inference. 

\subsection{Trajectory Prediction and Interactivity Estimation}

\begin{figure}[!tbp]
	\centering
	\includegraphics[width=\columnwidth]{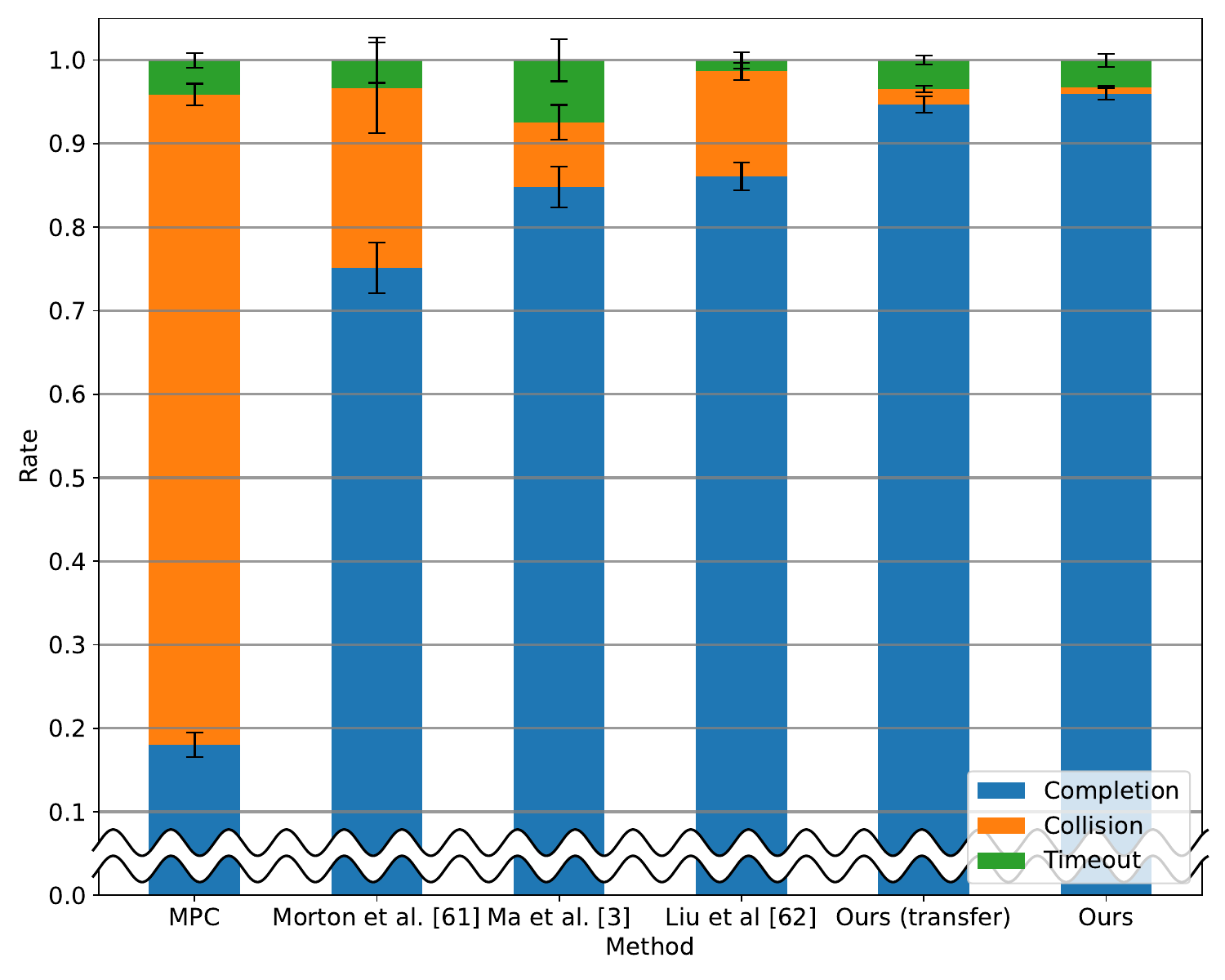}
	\caption{\rv{The comparison of the decision making performance in the simplified T-intersection scenario. Ours (transfer) means that the driving policy is trained in the four-way intersection and tested in the T-intersection. Ours means that the driving policy is both trained and tested in the T-intersection.}}
	\label{fig:baseline3}
\end{figure}

\begin{figure}[!tbp]
	\centering
	\includegraphics[width=\columnwidth]{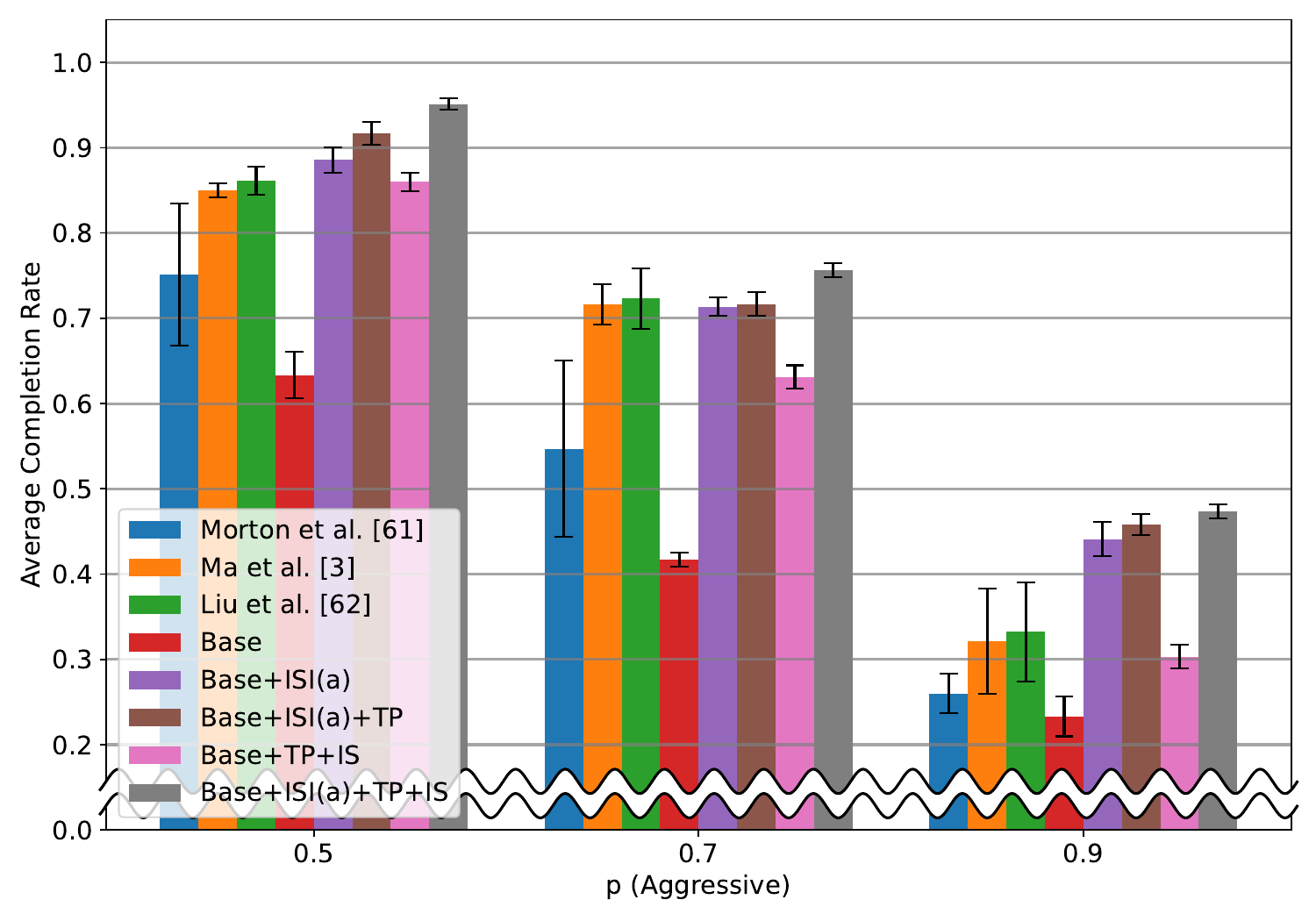}
	\caption{\rv{The comparison of average completion rates with different ratios of aggressive drivers in testing scenarios. A larger $p(\textsc{Aggressive})$ leads to more challenging scenarios. Note that we omit the bar plots in the small number range to better show the differences between methods.}}
        \vspace{-0.3cm}
	\label{fig:aggressiveness_shift}
\end{figure}

We demonstrate the effectiveness of trajectory prediction and interactivity estimation by comparing our complete framework Base+ISI(a)+TP+IS with its counterparts without the TP or IS modules.
Note that when TP is used alone without interactivity estimation, the $L_2$ losses of all the cases are equally weighted during training.
In Fig. \ref{fig:baselines}(a), we can observe an increase in the completion rate and a decrease in the collision rate and timeout rate by adding TP to Base+ISI(a). 
The comparison between Base+ISI(a) and Base+ISI(a)+TP in Fig. \ref{fig:baselines}(b) also shows that trajectory prediction also enhances trait and intention inference.
The reason is that the internal state of a surrounding vehicle can determine its future behavior, thus trajectory prediction can in turn encourage the model to capture subtle cues of the internal state implicitly.

Moreover, adding the interactivity estimation module leads to further improvements by providing the policy network about to what extent the ego vehicle can have potential influence on other agents through the interactivity scores. The implied quantitative degree of influence can facilitate the learning of negotiation.
The proposed method Base+ISI(a)+TP+IS achieves the best performance in Fig. \ref{fig:baselines}.
We found that most collisions are caused by wrong inference of internal states and inaccurate trajectory prediction of surrounding vehicles.

The comparison between Base+ISI(a)+TP(E)+IS and Base+ISI(a)+TP+IS implies the benefit of weighting the prediction error of each agent by its interactivity score in the loss function of trajectory prediction.
The reason is that, with the weighting mechanism, the predictor cares more about the prediction accuracy of the agents that may have strong interactions with the ego vehicle, which enhances ego decision making in challenging situations.

\rv{The comparison between our method and the state-of-the-art baselines \cite{ma2021reinforcement} (our prior work), \cite{morton2017simultaneous}, and \cite{liu2022learning} is shown in Fig. \ref{fig:baseline3}. Since the baseline methods cannot handle the pedestrians and the vehicles on the vertical lanes, we simplified the scenario into a T-intersection similar to the baseline papers for a fair comparison. Meanwhile, in these experiments, conservative vehicles always yield and aggressive ones do not yield. 
MPC method performs the worst due to its limited ability to handle diverse social behaviors and multi-agent interactions. It also requires online re-planning at a high frequency, which demands a large amount of computational resources. 
Our method achieves much better performance in terms of all the evaluation metrics. In particular, Ours (transfer) trained in the original environment (i.e., four-way intersection) achieves comparable performance with the one directly trained in the T-intersection, indicating good generalization performance.}

\begin{figure*}[!tbp]
	\centering
	\includegraphics[width=0.93\textwidth]{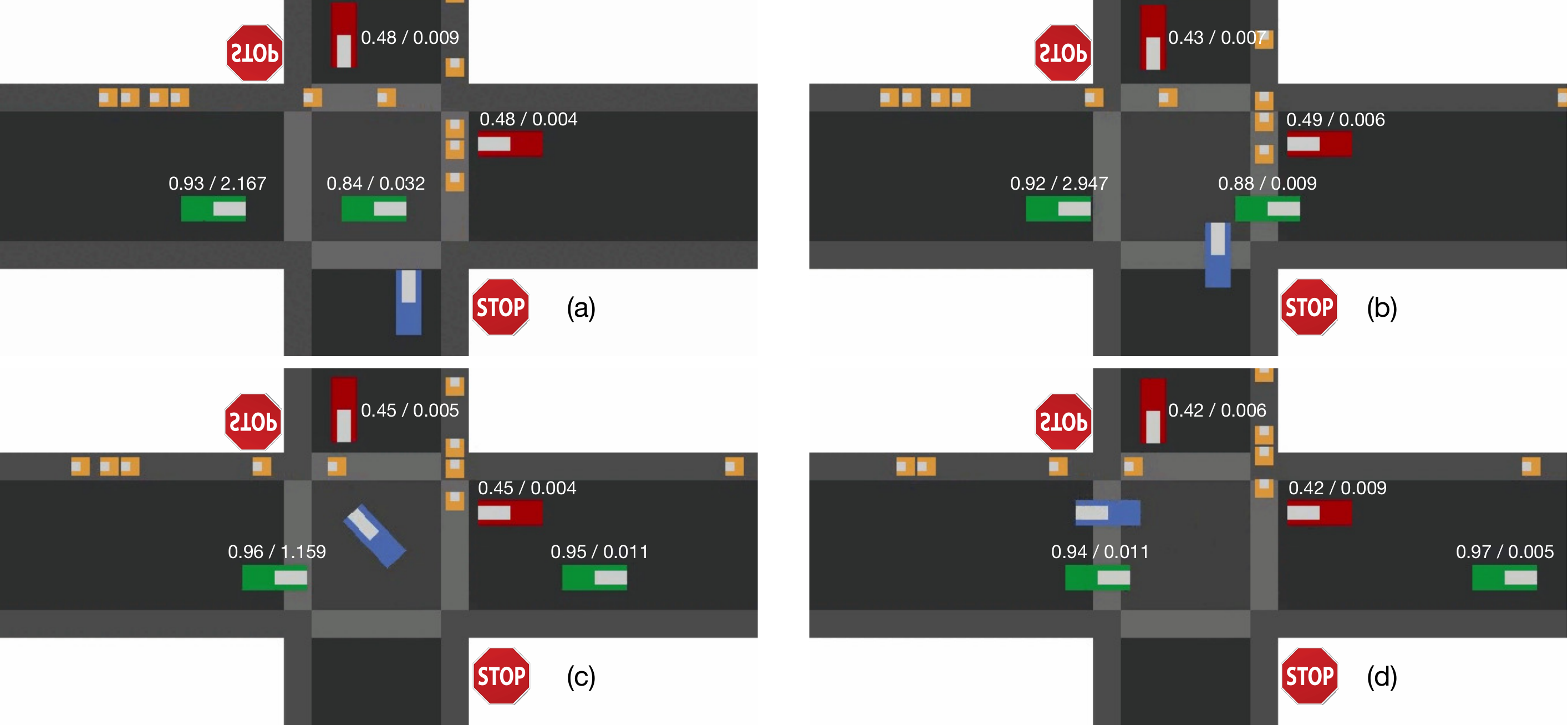}
	\caption{\rv{Four representative frames in a typical testing scenario. The gray rectangle inside each vehicle or pedestrian indicates its heading direction. The colors of surrounding vehicles indicate their ground truth traits. The first number near each vehicle denotes the inferred probability that the vehicle is a conservative one, and the second number denotes its interactivity score. Best viewed in color.}}
        \vspace{-0.3cm}
	\label{fig:visualization_scenarios}
\end{figure*}

\subsection{Robustness to Distribution Shift}
A critical aspect of a deep reinforcement learning method is its robustness to the distribution shift in the environments during the testing phase.
In the training phase, $p(\textsc{Aggressive})$ is fixed at 0.5 in our simulator.
To evaluate the robustness of different approaches, we changed $p(\textsc{Aggressive})$ to 70\% or 90\% for testing, which generates more aggressive drivers and leads to more challenging scenarios.

We have four observations based on the comparison of average completion rates shown in Fig. \ref{fig:aggressiveness_shift}.
First, the internal state inference shows great effectiveness, especially in environments with more aggressive drivers. The ablation models with ISI(a) improve the average completion rate by a large margin compared with their counterparts without ISI(a) in all settings.
The reason is that the ISI(a) module can recognize the internal characteristics of human drivers and infer their intentions explicitly, which helps the policy network to choose appropriate actions.
In environments with a large portion of aggressive drivers, it is necessary to accurately recognize the conservative ones so that the ego vehicle can grasp the opportunities to proceed safely and efficiently. 
Second, by comparing Base+ISI(a) and Base+ISI(a)+TP, we can see that trajectory prediction leads to improvement. The reason is that the additional supervision on trajectory prediction can improve internal state inference through learning better graph-based encoder since the trajectories can reflect the drivers' internal characteristics, which enhances decision making.
Third, a large portion of non-completion cases of the method settings without ISI(a) in the environment with $p(\textsc{Aggressive})=0.9$ is due to timeout caused by inefficient driving policies.
We can see an increasing trend of improvement brought by the interactivity estimation as the ratio of aggressive drivers increases, which implies the effectiveness of interactivity estimation in challenging scenarios.
Finally, the performance gaps between the baselines \cite{ma2021reinforcement,liu2022learning} and our method increase as more aggressive drivers appear in the scenarios, implying our method better handles challenging situations.

\subsection{\rv{Interpretation of the Decision Making Process}}\label{sec:explainability}

\rv{We provide the visualization of a typical testing scenario in Fig. \ref{fig:visualization_scenarios} to demonstrate the two aspects of the interpretation of the ego vehicle's decision making process more concretely.}
We select four representative frames for qualitative analysis, where the inferred internal states and interactivity scores are shown near each surrounding vehicle. 
In this work, we do not consider the internal states of pedestrians. Their interactivity scores are all very small since they always have the highest right of way, which are omitted in the figure for clarity.

In Fig. \ref{fig:visualization_scenarios}(a), the middle green vehicle is already in the conflict zone initially and the two trajectory prediction heads predict similar future trajectories, which leads to a low interactivity score. This is reasonable because it is unlikely to be influenced by the ego vehicle since it already occupies the conflict zone before the ego vehicle. 
However, the green one on the left has a much higher chance of being influenced by the ego vehicle, which is aligned with a high estimated interactivity score caused by more distinct trajectory hypotheses generated by the two prediction heads.
The method also infers these two vehicles as conservative ones confidently based on their historical behavior patterns.

Due to the existence of the crossing pedestrians, the two red aggressive vehicles need to stop and wait until their paths are clear. This provides the ego vehicle a good opportunity to turn left safely without negotiating with the red vehicle on the main road.
In this situation, the red vehicles have low interactivity scores because their near-future trajectories will not change no matter whether the ego vehicle exists or not due to the constraints caused by the crossing pedestrians.
Note that they have non-zero interactivity scores because of prediction errors.
This implies that our method successfully learns the underlying relations between vehicles and pedestrians. 
In Fig. \ref{fig:visualization_scenarios}(b)(c), the low interactivity scores of red vehicles combined with the physical state information informs the policy that the ego vehicle can proceed safely without a strong negotiation even though the opponent is an aggressive vehicle.
In Fig. \ref{fig:visualization_scenarios}(d), the interactivity score of the left green vehicle becomes much smaller because the ego vehicle no longer influences its future behavior after crossing the intersection.

\subsection{Influence of Pedestrians}

\begin{figure*}[!tbp]
	\centering
	\includegraphics[width=\textwidth]{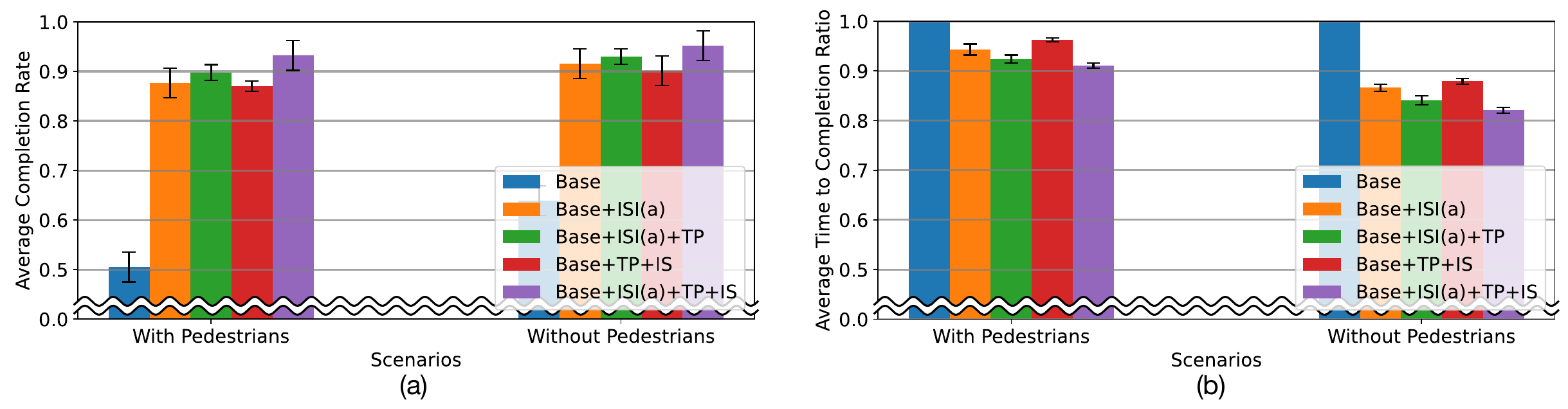}
	\caption{\rv{The influence of pedestrians on the decision making performance. The policy trained in environments with pedestrians is also tested in environments without pedestrians. (a) Average completion rate; (b) Average time-to-completion ratio: the ratio is calculated with respect to the Base method. }}
	\label{fig:influence_peds}
\end{figure*}

In Fig. \ref{fig:visualization_scenarios}, a typical testing scenario is visualized to qualitatively illustrate the influence of crossing pedestrians on the vehicles, where the vehicles on the main road need to yield to the crossing pedestrians thus the ego vehicle is able to proceed.
To quantitatively show the influence of crossing pedestrians on the decision making of the ego vehicle, we compare the performance in the environments with pedestrians and without pedestrians in Fig. \ref{fig:influence_peds}.
In general, the average completion rate is higher and the average time-to-completion is shorter in environments without pedestrians for all the methods. This is reasonable because the existence of crossing pedestrians leads to more challenging situations where the ego vehicle needs to yield to the pedestrians as well as seize the opportunity to make the left turn as fast as possible.
The results show consistent relative performance among different methods. 

\section{Conclusions}
In this paper, we present a deep reinforcement learning framework with auxiliary supervised learning tasks for autonomous navigation and use a partially controlled intersection scenario as a case study to validate our method.
First, we propose to infer the internal state of surrounding human-driven vehicles including their traits (i.e., conservative/aggressive) and intentions (i.e., yield/not yield to the ego vehicle).
Second, we propose to estimate the interactivity scores of traffic participants based on the inferred degree of influence via counterfactual trajectory prediction to provide additional cues to the policy network.
These auxiliary tasks improve the completion rate, reduce collisions, and enhance driving efficiency by a large margin compared with state-of-the-art baselines.
The ablation study demonstrates the effectiveness of each component of our method.
In particular, our method is more robust to the distribution shifts in the testing environments and provides explainable intermediate indicators for ego decision making.
Moreover, we design an intersection driving simulator based on an Intelligent Intersection Driver Model, which is used to simulate the interactive behaviors of vehicles and pedestrians in our experiments.

The limitation of this work lies in the gap between driving simulation and real-world scenarios.
In this work, we assume that human drivers can be divided into binary groups (i.e., conservative or aggressive), which is a reasonable simplification of human traits. Our method requires ground truth labels for human traits and intentions provided by the simulator in the training process.
However, human traits could be more complicated in the real world and the ground truth labels may not be straightforward to obtain.
The goal of this study is to validate the hypothesis that modeling human internal states explicitly can improve decision making performance and the inferred internal states can serve as explainable indicators.
In future work, we plan to address this limitation by learning human traits and intentions with latent variable models in an unsupervised manner, which eliminates the demand for ground truth labels and allows for modeling more flexible internal states than binary categories.
\rv{We will also investigate how to guarantee safety in a principled manner.}


\section{Appendix}\label{sec:appendix}

In this section, we introduce the detailed operations in other message passing mechanisms (besides GAT) used in our experiments and analyze the experimental results.

\subsection{Graph Message Passing Mechanisms}

\subsubsection{GCN}
This model applies convolution operations to graphs. We formulate a node attribute matrix $\mathbf{V}$ where each row denotes the attribute of a certain node. Then the updated node attribute matrix $\bar{\mathbf{V}}$ can be obtained by
		\begin{align}
			\bar{\mathbf{V}}_t = \sigma \left( \tilde{\mathbf{D}}^{-\frac{1}{2}} \tilde{\mathbf{A}} \tilde{\mathbf{D}}^{-\frac{1}{2}} \mathbf{V}_t \mathbf{W} \right),
		\end{align}
	where $\tilde{\mathbf{A}} = \mathbf{A} + \mathbf{I}$ is the adjacency matrix of $\mathcal{G}_t$ with self-connections. $\mathbf{I}$ is the identity matrix. $\tilde{\mathbf{D}}^{ii} = \sum_{j} \tilde{\mathbf{A}}^{ij}$ and $\mathbf{W}$ is a learnable weight matrix. Here, $\sigma(\cdot)$ denotes a nonlinear activation function.

 \subsubsection{GraphSAGE}
This model designs a customized message passing mechanism, which includes the following operations:
\begin{align}
    \text{MSG}^i_t =& \ f_\text{AGG} \left( \left\{ \mathbf{v}^j \mid \forall j \in \mathcal{N}^i \right\} \right),\\
    \bar{\mathbf{v}}^i_t =& \ \sigma \left( \mathbf{W} \left[ \mathbf{v}^i_t \| \text{MSG}^i_t \right] \right),\\
    \bar{\mathbf{v}}^i_t \leftarrow& \ \bar{\mathbf{v}}^i_t / \|\bar{\mathbf{v}}^i_t\|_2,
\end{align}
where $\text{MSG}^i_t$ denotes an intermediate message obtained by aggregating the information from the neighbors of node $i$, and $f_\text{AGG}$ is an arbitrary permutation invariant function. $\mathbf{W}$ denotes a learnable weight matrix and $\sigma(\cdot)$ denotes a nonlinear activation function.

\subsection{Influence of GNN Architectures}

\begin{figure}[!tbp]
	\centering
	\includegraphics[width=\columnwidth]{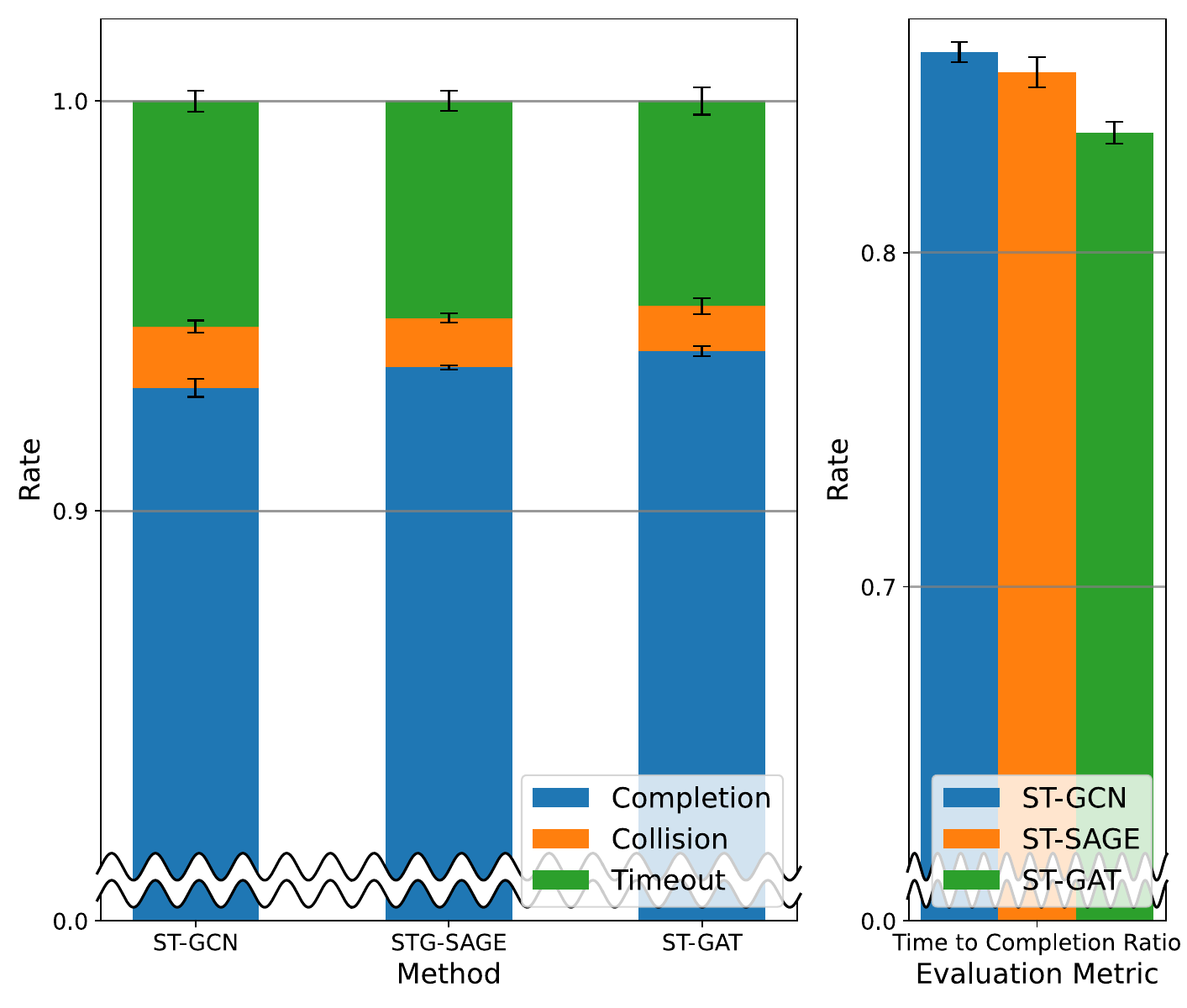}
	\caption{\rv{The comparison of average completion rates with different variants of spatio-temporal graph neural networks in the graph-based encoder. The time-to-completion ratio is computed in the same way as in Fig. \ref{fig:baselines}.}}
    \vspace{-0.3cm}
	\label{fig:GNN_variants}
\end{figure}

We conducted an ablation study on several widely used graph neural network architectures for spatio-temporal graph modeling: ST-GAT \cite{huang2019stgat}, ST-SAGE \cite{ma2021reinforcement}, and ST-GCN \cite{yu2018spatio}. The comparison of results is shown in Fig. \ref{fig:GNN_variants}. Generally, there is no significant gap in performance with different architectures of graph neural networks, which implies that our method is not sensitive to the choice of graph neural networks.

ST-GCN is a modified graph convolutional network that is applied to spatio-temporal graphs, which uses the graph adjacency matrix and applies simple convolutions across the graph.
ST-SAGE has a more expressive message passing mechanism than ST-GCN and leverages the node attribute information more effectively, which leads to better performance.
ST-GAT applies a graph attention mechanism in message passing and achieves the best performance.
A potential reason is that the graph attention layers naturally learn to recognize and use important information in the node updates, which is suitable for gathering information from other agents.

{
\bibliographystyle{IEEEtran}
\bibliography{references}
}

\begin{IEEEbiography}[{\includegraphics[width=1in,height=1.25in,clip,keepaspectratio]{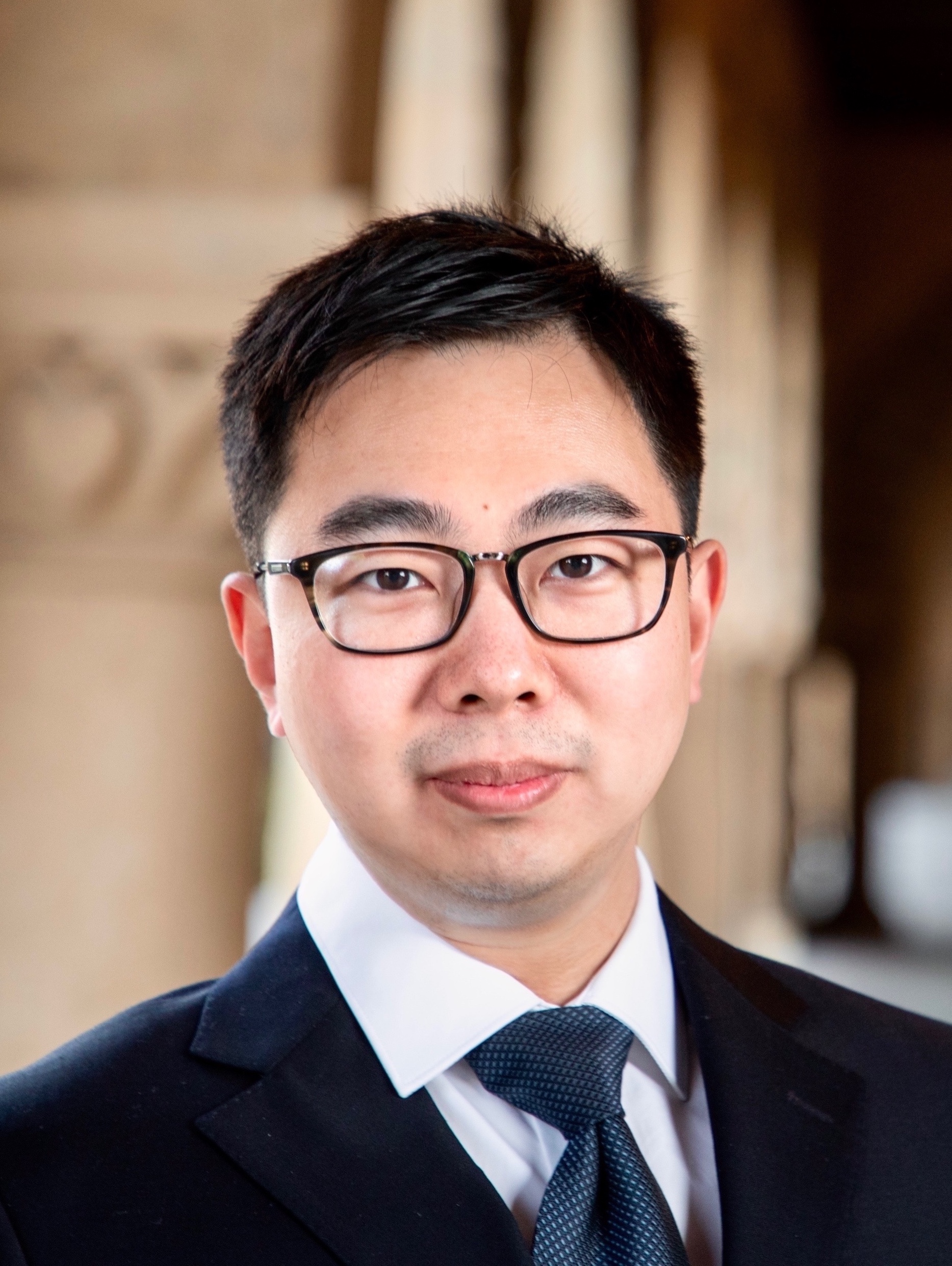}}]{Jiachen Li}
(Member, IEEE) received his Ph.D. degree from the Department of Mechanical Engineering at the University of California, Berkeley in 2021. Before that, he received a B.E. degree from the Department of Control Science and Engineering at Harbin Institute of Technology, China in 2016. He is currently a postdoctoral scholar at Stanford University. His research interest lies at the broad intersection of robotics, trustworthy AI, reinforcement learning, control and optimization, and their applications to intelligent autonomous systems, particularly in human-robot interactions and multi-agent systems. Dr. Li was selected as an RSS Robotics Pioneer in 2022 and an ASME DSCD Rising Star in 2023. He serves as an associate editor or a reviewer for multiple journals and conferences. He has organized multiple workshops on robotics, machine learning, computer vision, and intelligent transportation systems. 
\end{IEEEbiography}

\begin{IEEEbiography}[{\includegraphics[width=1in,height=1.25in,clip,keepaspectratio]{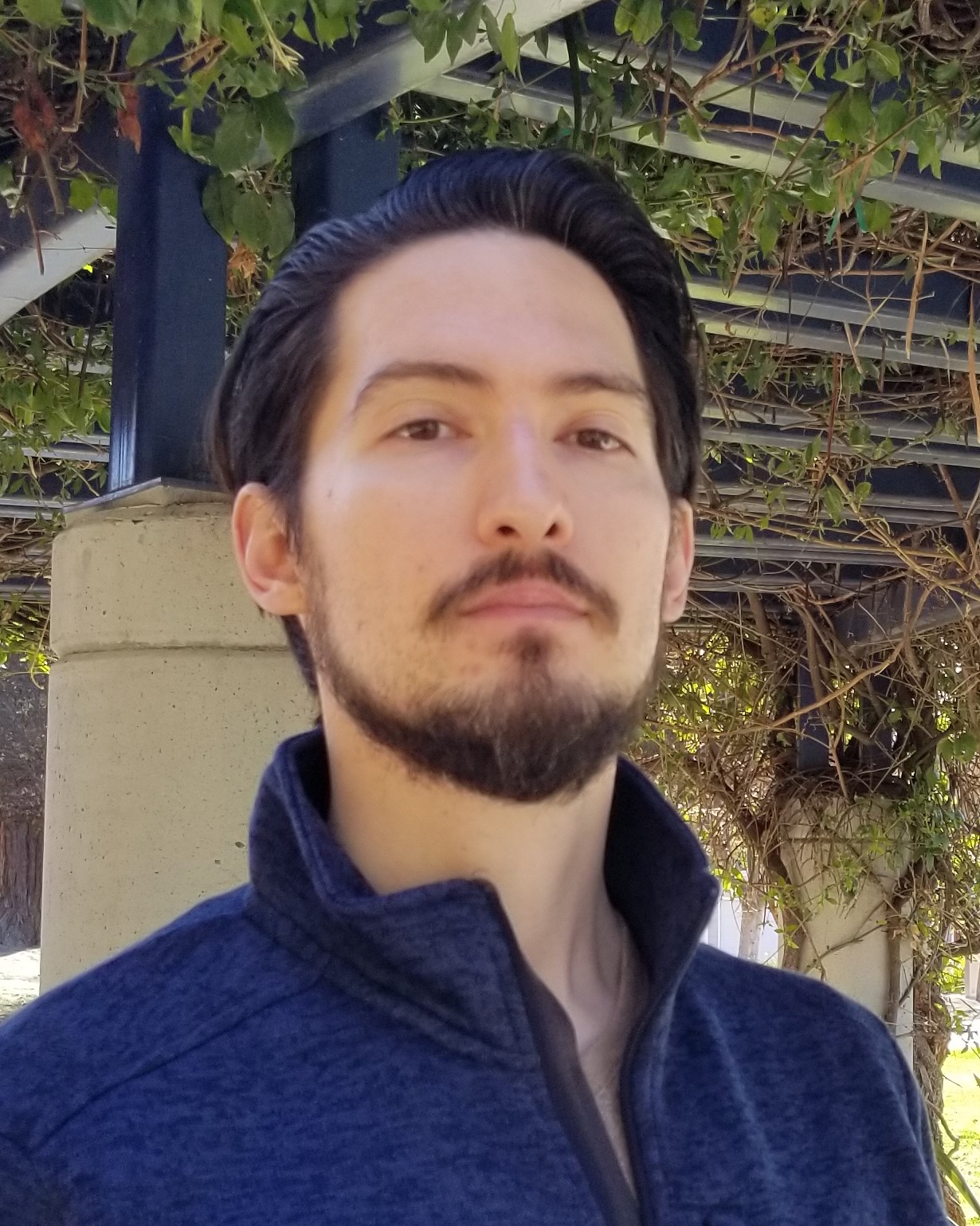}}]{David Isele} (Member, IEEE) received his M.S.E. degree in Robotics and his Ph.D. degree in Computer and Information Science from The University of Pennsylvania. Before that, he received his B.E. degree in electrical engineering from The Cooper Union: Albert Nerken School of Engineering, in New York City.
He is currently a senior scientist at Honda Research Institute US. His research interests include applications of machine learning and artificial intelligence to robotic systems, with a focus on strategic decision making for autonomous vehicles. 
\end{IEEEbiography}

\begin{IEEEbiography}[{\includegraphics[width=1in,height=1.25in,clip,keepaspectratio]{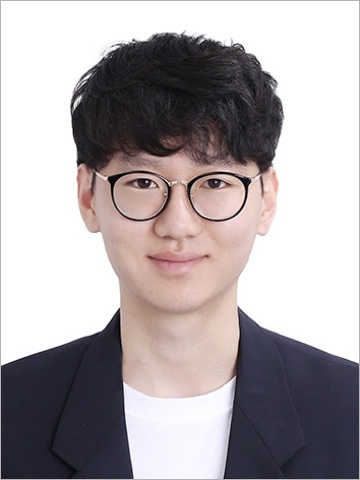}}]{Kanghoon Lee} received his BS degree in Industrial and Systems Engineering and Computer Science from the Korea Advanced Institute of Science and Technology (KAIST), South Korea, in 2020, and MS degree in Industrial and Systems Engineering from the KAIST, South Korea, in 2022.

Currently, He is a Ph.D. candidate in the System Intelligence Laboratory at the Department of Industrial and Systems Engineering, KAIST, South Korea. His research interest lies at the intersection of machine learning, (multi-agent) reinforcement learning and their application to robotic or traffic systems.
\end{IEEEbiography}

\begin{IEEEbiography}[{\includegraphics[width=1in,height=1.25in, clip,keepaspectratio]{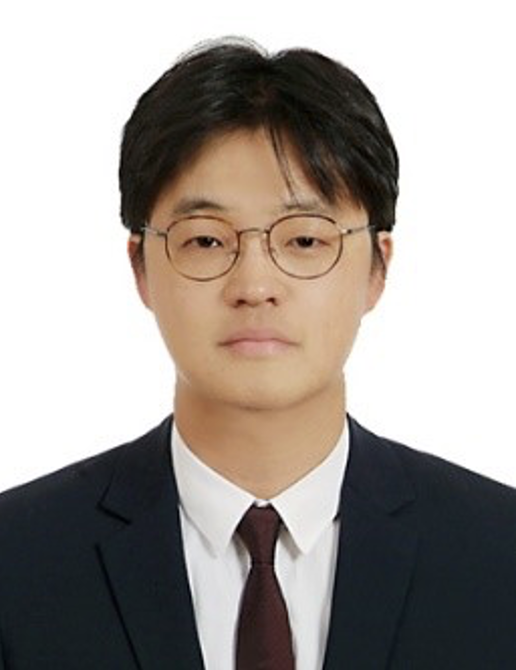}}]{Jinkyoo Park} is currently an associate professor of Industrial and Systems Engineering and adjoint professor of Graduate School of Artificial Intelligence at Korea Advanced Institute of Science and Technology (KAIST), Republic of Korea. He received his B.S. degree in Civil and Architectural Engineering from Seoul National University in 2009, an M.S. degree in Civil, Architectural and Environmental Engineering from the University of Texas Austin in 2011, an M.S. degree in Electrical Engineering from Stanford University in 2015, and a Ph.D. degree in Civil and Environmental Engineering from Stanford University in 2016. His research goal is to explore the potential of the various machine learning approaches for improving complex decision-making methods in optimization, optimal control, and game theory.
\end{IEEEbiography}

\begin{IEEEbiography}[{\includegraphics[width=1in,height=1.25in,clip,keepaspectratio]{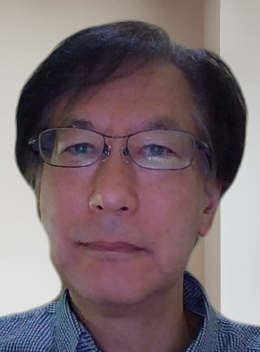}}]{Kikuo Fujimura}
received the B.S. and M.S. degrees in Information Science from the University of Tokyo in 1983 and 1985, respectively, and the Ph.D. degree in Computer Science from the University of Maryland, College Park in 1989.  After working at Oak Ridge National Laboratory and Ohio State University (Columbus), he joined Honda R\&D in 1998, where he was engaged in research on intelligent systems and human-robot interaction with Honda’s humanoid robot ASIMO. He is currently Director of Innovation at Honda Research Institute USA in San Jose, California, where he directs teams of researchers working on automated driving, knowledge discovery and informatics, human-machine interfaces, and intelligent robotics.  

His research interests include artificial intelligence for mobility, human-robot interaction, and HCI. He has authored/co-authored one book and over 100 research papers in refereed conferences and journals and has been granted over 20 patents. He currently serves as an Associate Editor for IEEE Transactions on Intelligent Vehicles and ACM Journal on Autonomous Transportation Systems. 

\end{IEEEbiography}

\begin{IEEEbiography}[{\includegraphics[width=1in,height=1.25in,clip,keepaspectratio]{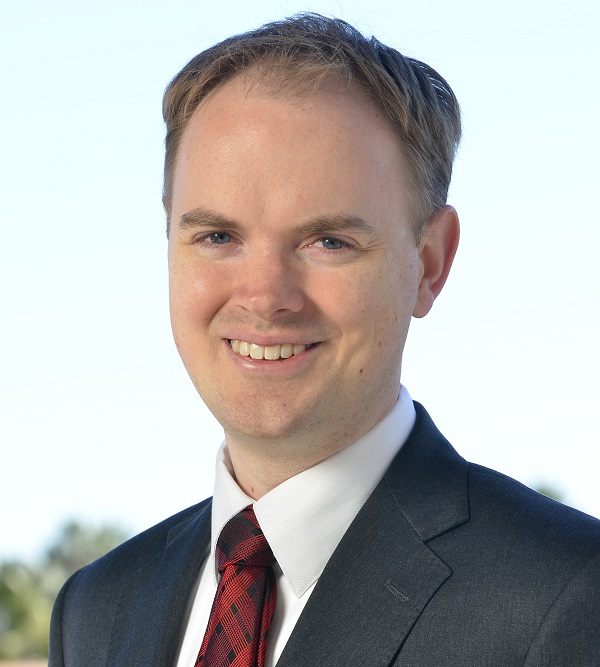}}]{Mykel J. Kochenderfer}
(Senior Member, IEEE) is an Associate Professor of Aeronautics and Astronautics at Stanford University. He is the director of the Stanford Intelligent Systems Laboratory (SISL), conducting research on advanced algorithms and analytical methods for the design of robust decision making systems. Prior to joining the faculty in 2013, he was at MIT Lincoln Laboratory where he worked on aircraft collision avoidance for manned and unmanned aircraft. He received his Ph.D. from the University of Edinburgh in 2006. He received B.S. and M.S. degrees in computer science from Stanford University in 2003. He is an author of the textbooks \textit{Decision Making under Uncertainty: Theory and Application} (MIT Press, 2015), \textit{Algorithms for Optimization} (MIT Press, 2019), and \textit{Algorithms for Decision Making} (MIT Press, 2022).

\end{IEEEbiography}
\vfill

\end{document}